\documentclass{article} % For LaTeX2e
\usepackage[final]{colm2026_conference}

\usepackage{microtype}
\usepackage{hyperref}
\usepackage{url}
\usepackage{booktabs}
\usepackage{amsmath, amsthm}
\usepackage{algorithm, algorithmic}
\usepackage{color} 
\usepackage{graphicx}
\usepackage{longtable, array}

\usepackage{lineno}

\definecolor{darkblue}{rgb}{0, 0, 0.5}
\hypersetup{colorlinks=true, citecolor=darkblue, linkcolor=darkblue, urlcolor=darkblue}
\newcommand{\method}{\texorpdfstring{\mbox{\textsc{SL}\textsuperscript{2}}}{SL2}}

\usepackage{amsmath,amsfonts,bm}

\def\eqref#1{equation~\ref{#1}}
\def\1{\bm{1}}

\DeclareMathAlphabet{\mathsfit}{\encodingdefault}{\sfdefault}{m}{sl}
\SetMathAlphabet{\mathsfit}{bold}{\encodingdefault}{\sfdefault}{bx}{n}

\title{Spend Less, Fit Better: Budget-Efficient Scaling Law Fitting via Active Experiment Selection}

\author{Sijie Li$^{1}$\thanks{Equal contribution.}, ~Shanda Li$^{2}$\footnotemark[1], ~Haowei Lin$^{1}$, Weiwei Sun$^{2}$, Ameet Talwalkar$^{2,3}$, Yiming Yang$^{2}$\\[0.5em]
$^1$Peking University\quad
$^2$Carnegie Mellon University\quad
$^3$Datadog\\
}

\begin{document}

\ifcolmsubmission
\linenumbers
\fi

\maketitle

\begin{abstract}
Scaling laws are used to plan multi-million-dollar training runs, but fitting those laws can itself cost millions. In modern large-scale workflows, assembling a sufficiently informative set of pilot experiments is already a major budget-allocation problem rather than a routine preprocessing step. We formulate scaling-law fitting as budget-aware sequential experimental design: given a finite pool of runnable experiments with heterogeneous costs, choose which runs to execute so as to maximize extrapolation accuracy in a high-cost target region. We propose \method{} (Scaling Laws, Spend Less), an uncertainty-aware method for sequentially allocating experimental budget toward the runs most useful for target-region extrapolation. Across a diverse benchmark of scaling-law tasks, \method{} outperforms classical design-based baselines, and often approaches the performance of fitting on the full experimental set while using only about 10\% of the total training budget. Our code is available at \url{https://github.com/PlanarG/active-sl}.
\end{abstract}

\section{Introduction}

Scaling laws have become a central tool for analyzing and planning large-scale language model training \citep{hestness2017deep, kaplan2020scaling, gao2023scaling, li2025predictablescaleiifarseer}. By fitting a parametric relationship between performance and variables such as model size, data size, and compute, practitioners can use a limited set of pilot runs to predict behavior at much larger scales and allocate future training budget accordingly \citep{yang2022tensor,li2025predictable, xu2026width}. This paradigm has already shaped influential decisions in practice: For example, Chinchilla-style compute-optimal training was derived from an extensive empirical study spanning more than 400 training runs across a wide range of scales \citep{hoffmann2022trainingcomputeoptimallargelanguage}.

Despite their practical importance, scaling laws remain expensive to fit and heavily reliant on manual experiment design. In typical workflows, researchers hand-select experimental configurations, run many pilot trainings, and then fit a parametric law to the resulting observations. At industrial scale, the pilot runs needed just to fit a scaling law can themselves consume a massive budget \citep{porian2024resolving, hagele2024scaling}, with the full fit-and-verify pipeline already reaching the \textbf{million-dollar scale} before any deployment-scale training is committed. Accurate scaling-law fitting is therefore not only a modeling problem, but also a problem of budget allocation.

\begin{figure}[t]
    \centering 
    \includegraphics[width=\linewidth]{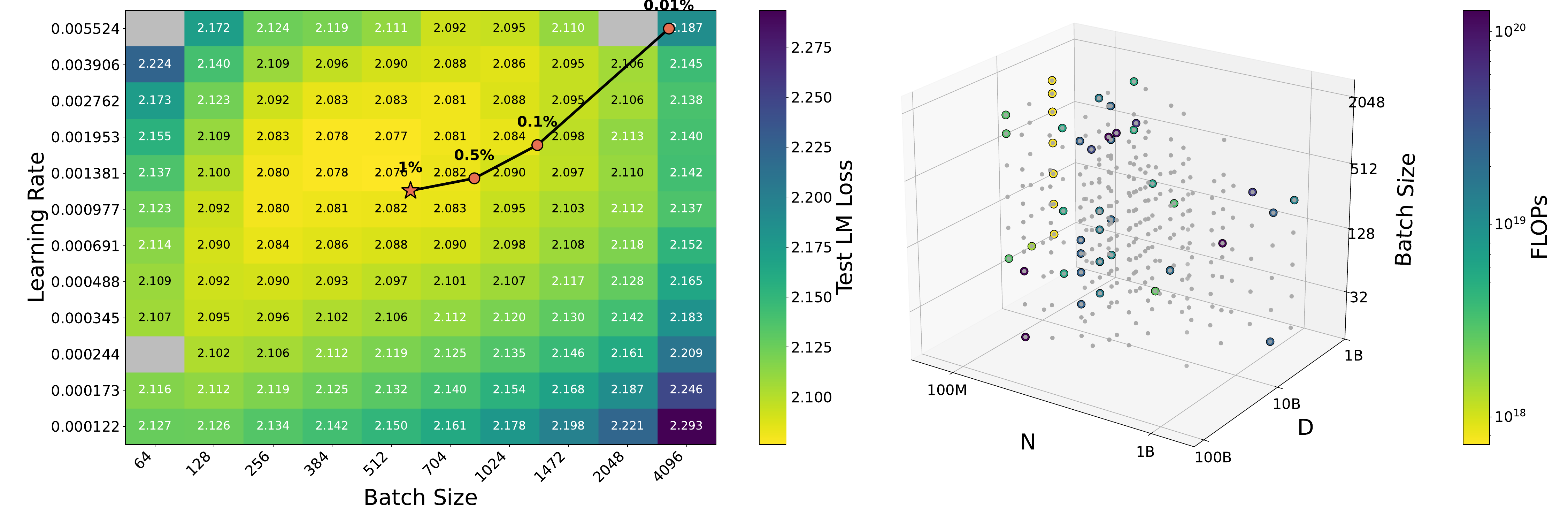}
    \vspace{-6mm}
    \caption{\textbf{Budget-efficient extrapolation} on \texttt{lr\&bsz}. \method{} identifies the extrapolation optimum using only a small fraction of the original scaling-law fitting budget. On \texttt{lr\&bsz}, the predicted optimum over learning rate and batch size for a 1B-parameter LLM trained on 100B tokens reaches the low-loss region within 1\% of the budget, as shown by the trajectory on the test-loss heatmap (left). The selected training configurations used for fitting are shown on the right in a 3D view of the design space, colored by training cost ($6ND$), illustrating that accurate extrapolation can be achieved from a sparse, low-cost subset of the full configuration space.}
    \label{fig:teaser}
\end{figure}

This challenge is becoming more acute as scaling-law analysis expands beyond classical dense pretraining. Recent work has extended scaling laws to a much broader range of settings, including vocabulary design, data mixing, sparsity, mixture-of-experts architectures, and inference-time scaling \citep{muennighoff2023scaling, caballero2023brokenneuralscalinglaws, krajewski2024scalinglawsfinegrainedmixture}. As these settings become more diverse and expensive, manually selecting pilot runs becomes increasingly inefficient. This raises a natural question: 
\begin{center}
    \textit{Given a pool of runnable experiments and a limited budget, how should we select experiments to ensure that the fitted scaling law extrapolates accurately in the target region?}
\end{center}

In this paper, we formulate scaling-law fitting as a problem of \emph{budget-aware sequential experimental design}. Rather than assuming that fitting data are given in advance, we treat each candidate experiment as a costly query and ask how to allocate a limited budget over a finite pool of runnable configurations. The objective is not merely to fit the observed points well, but to maximize predictive accuracy in a held-out target region lying in the large-scale, high-cost regime. This formulation more directly reflects the practical use of scaling laws, where the ultimate goal is to choose a few expensive final configurations rather than to uniformly fit all pilot runs. To address this objective, we propose \method{} (Scaling Laws, Spend Less), a sequential, uncertainty-aware design method for budget-efficient scaling-law fitting.

Our contributions are twofold. 
\begin{itemize}
    \item We formulate the problem and present the first systematic study of scaling-law fitting in the low-budget regime, where accurate extrapolation must be recovered from only a small fraction of the training runs used in standard scaling-law analyses. To support this study, we conduct a comprehensive evaluation spanning diverse scaling scenarios, heterogeneous law families, and task-specific cost structures, enabling controlled comparison of point-selection strategies under budget constraints.
    \item \method{} explicitly models ambiguity over scaling-law parameters and selects new experiments according to their expected value for reducing target-region prediction error. This positions scaling-law fitting as a target-extrapolation compute-allocation problem rather than a standard parameter-estimation design problem. Empirically, our method is highly effective: as shown in Figure~\ref{fig:teaser}, it approaches the extrapolation ground-truth optimum using only a small fraction of the original fitting FLOPs, reaching the low-loss region within $1\%$ budget on \textit{Step Law} (Learning Rate and Batch Size Scaling Law) \citep{li2025predictable}.
\end{itemize}

\section{Related Work}

\paragraph{Scaling Laws.} Scaling laws have transformed the design and optimization of large-scale AI systems by revealing predictable relationships among model size, data volume, and compute budget \citep{cortes1993learning,kaplan2020scaling, hoffmann2022trainingcomputeoptimallargelanguage, hu2024minicpm}. Recent work has extended this paradigm far beyond its original setting, covering model architectures \citep{fedus2021switch}, data scaling \citep{ye2025data, que2024dcpt}, post-training behavior \citep{hernandez2021transfer, lin2024selecting}, multimodal regimes \citep{radford2021learning, zhai2022scalingvisiontransformers}, and inference-time settings \citep{chen2024are, chen2025parallelscalinglawlanguage, wu2025inference,huang2026sample}. Yet fitting scaling laws in practice remains costly and heavily manual: the final fit can be highly sensitive to the choice of law family, initialization, and, crucially, the strategy used to collect training runs \citep{li2025misfittingsurveyscalinglaws}. \method{} automatically selects new runs based on the current fitting state and the target region where accurate prediction matters most, without relying on manually designed experimental points.

\paragraph{Optimal experiment design for nonlinear models.} 

Classical optimal experiment design (OED) studies how to place experiments to estimate model parameters or derived quantities efficiently, with criteria such as D-optimality and A-optimality defined through the Fisher information matrix \citep{silvey2013optimal, kiefer1959optimum, gilmour2012optimum}. For nonlinear models, these criteria typically depend on unknown parameters, leading to a large literature on locally optimal design for nonlinear and generalized linear models \citep{white1973extension, khuri2006design, atkinson2015designs, yang2013optimal}. However, this line of work is primarily local and parameter-estimation-oriented, and typically does not consider heterogeneous experiment costs. Bayesian optimal experimental design addresses parameter uncertainty by optimizing expected utility under a posterior distribution \citep{huan2013simulatio, zhong2026goaloriented}, which is appealing for scaling-law fitting where the objective is highly nonlinear and limited observations may support multiple plausible fits. However, existing Bayesian OED methods do not directly address our setting, in which candidate experiments are discrete training runs with heterogeneous compute costs and must be selected sequentially under a strict budget. We therefore study a cost-aware sequential design problem tailored to scaling-law fitting, where experiments are prioritized by their expected predictive benefit relative to their cost.

\section{Problem Setup}

We study the scaling-law fitting problem in a budget-constrained sequential setting. We assume that the underlying performance trend is described by a parametric scaling law $y=f(x;\theta)$, where $x \in \mathcal X$ denotes the modeling configuration, e.g., model size, token count, or other training- or inference-related hyperparameters; $y \in \mathbb R$ denotes the prediction target of the scaling law, e.g., training loss; and $\theta \in \mathbb R^p$ denotes the parameter to be estimated from the experiments. Running an experiment under configuration $x$ incurs a nonnegative cost $c(x)$ and reveals an outcome $y$. At a high level, we aim to run a set of experiments under a cost constraint so as to obtain data for fitting an accurate scaling law.

In practice, experiment selection is performed over a set of predefined runnable configurations. We therefore consider a candidate pool $\mathcal X_{\mathrm{cand}}=\{x_1,\dots,x_N\}$, where each candidate $x_i$ has an associated cost $c_i=c(x_i)$. At each round, the learner selects one previously unobserved candidate, pays its cost, observes its outcome, and adds the resulting pair to the current dataset. After $t$ rounds, the accumulated observations form a dataset $\mathcal D_t$, and the total cost of all selected experiments must remain within a budget $C$.

The ultimate goal is not merely to fit the observed points well, but to learn a scaling law that extrapolates accurately in a target region $\mathcal X_{\mathrm{tar}}$. This target region typically contains the larger-scale configurations that matter most for downstream planning, but are too expensive to explore exhaustively, whereas $\mathcal X_{\mathrm{cand}}$ typically contains more affordable small-scale runs. Our objective is therefore to design a sequential experiment-selection strategy that uses the available budget as efficiently as possible, so that the fitted scaling law is accurate where it ultimately matters.
\section{\method: Budget-Aware Sequential Scaling-Law Design}

We now describe \method, our sequential design strategy for scaling-law fitting under a budget constraint. At round $t$, given the current dataset $\mathcal D_t$, our goal is to select the next experiment $x \in \mathcal X_{\mathrm{cand}}$ that most improves predictive accuracy on the target region $\mathcal X_{\mathrm{tar}}$.

To make the design objective concrete, we assume the standard observation model $y = f(x;\theta^*) + \varepsilon$ with some unknown ground-truth parameter $\theta^*$ and noise $\varepsilon \sim \mathcal N(0,\sigma^2)$.

\subsection{A Target-Aware Uncertainty Objective}

We would like the utility of an experiment to reflect our downstream goal: improving prediction accuracy on the target region $\mathcal X_{\mathrm{tar}}$. In our setting, uncertainty comes from two sources: local uncertainty within a plausible fit, and disagreement across multiple plausible fits that extrapolate differently on $\mathcal X_{\mathrm{tar}}$. We call each locally optimal fit of the scaling law a \emph{\textbf{basin}}. Given the observations $\mathcal D_t$ collected so far, we approximate the posterior:
\[
p(\theta \mid \mathcal D_t)
\approx
\sum_{k=1}^K w_k\, q_k(\theta),
\qquad
q_k(\theta)=\mathcal N(\theta_k,\Sigma_k),
\]
where each component represents one plausible local basin, with representative parameter $\theta_k$, local covariance $\Sigma_k$, and mixture weight $w_k \ge 0$ satisfying $\sum_{k=1}^K w_k = 1$. In our implementation, $\theta_k$, $\Sigma_k$, and $w_k$ are estimated from local refits and local posterior approximations; details are given in Appendix~\ref{app:basin_estimation}.

To measure uncertainty where extrapolation matters, define the target-region prediction map
\[
F(\theta)
=
\bigl(f(x;\theta)\bigr)_{x\in\mathcal X_{\mathrm{tar}}}
\in \mathbb R^{|\mathcal X_{\mathrm{tar}}|}.
\]
Let
\[
\hat f_k = F(\theta_k),
\qquad
\bar f
=
\mathbb E_{\theta\sim p(\theta\mid\mathcal D_t)}[F(\theta)]
\approx
\sum_{k=1}^K w_k \hat f_k .
\]
We use the target-region mean squared prediction error
\[
\mathrm{MSPE}_{\mathrm{tar}}
=
\frac{1}{|\mathcal X_{\mathrm{tar}}|}
\mathbb E_{\theta\sim p(\theta\mid\mathcal D_t)}
\!\left[
\|F(\theta)-\bar f\|_2^2
\right]
\]
as our uncertainty objective. Under a local Gaussian approximation within each basin, this quantity decomposes into
\[
\mathrm{MSPE}_{\mathrm{tar}}
=
V_{\mathrm{intra}} + V_{\mathrm{inter}},
\]
where
\[
V_{\mathrm{intra}}
=
\frac{1}{|\mathcal X_{\mathrm{tar}}|}
\sum_{k=1}^K
w_k\,\operatorname{tr}(J_k \Sigma_k J_k^\top),
\qquad
V_{\mathrm{inter}}
=
\frac{1}{|\mathcal X_{\mathrm{tar}}|}
\sum_{k=1}^K
w_k\,\|\hat f_k-\bar f\|_2^2.
\]
Here $J_k \in \mathbb R^{|\mathcal X_{\mathrm{tar}}|\times p}$ is the Jacobian of $F(\theta)$ evaluated at $\theta_k$. Intuitively, the first term quantifies local predictive uncertainty within each basin, while the second quantifies disagreement across basins in the target region.

\subsection{Scoring Candidate Experiments}

For a candidate experiment $x \in \mathcal X_{\mathrm{cand}}$, we define its utility as the expected reduction in target-region uncertainty after observing its outcome:
\[
\Delta \mathrm{MSPE}_{\mathrm{tar}}(x)
=
\mathrm{MSPE}_{\mathrm{tar}}
-
\mathbb E_{y \mid x,\mathcal D_t}
\bigl[
\mathrm{MSPE}_{\mathrm{tar}}^{+}(x,y)
\bigr],
\]
where $\mathrm{MSPE}_{\mathrm{tar}}^{+}(x,y)$ denotes the updated target-region MSPE after augmenting $\mathcal D_t$ with $(x,y)$. Using the decomposition above, we write
\[
\Delta \mathrm{MSPE}_{\mathrm{tar}}(x)
=
\Delta V_{\mathrm{intra}}(x)
+
\Delta V_{\mathrm{inter}}(x),
\]
with
\[
\Delta V_{\mathrm{intra}}(x)
=
V_{\mathrm{intra}}
-
\mathbb E_{y \mid x,\mathcal D_t}
\bigl[
V_{\mathrm{intra}}^{+}(x,y)
\bigr],
\qquad
\Delta V_{\mathrm{inter}}(x)
=
V_{\mathrm{inter}}
-
\mathbb E_{y \mid x,\mathcal D_t}
\bigl[
V_{\mathrm{inter}}^{+}(x,y)
\bigr].
\]
The first term favors candidates that reduce within-basin predictive variance on $\mathcal X_{\mathrm{tar}}$, while the second favors candidates that disambiguate basins with different extrapolations. To account for heterogeneous experiment costs, we rank candidates by the cost-aware score
\[
S(x)
=
\frac{
\Delta V_{\mathrm{intra}}(x)
+
\Delta V_{\mathrm{inter}}(x)
}{
c(x)^\alpha
},
\]
where $\alpha \ge 0$ controls the strength of cost penalization.

\subsection{Computing Intra- and Inter-Basin Utility}

We approximate both utility terms by locally linearizing the predictor within each basin while preserving the multimodal structure across basins. Full derivations are deferred to Appendix~\ref{app:utility_derivation}.

\paragraph{Intra-basin utility.}
For a candidate $x$, let
\[
J_x(\theta_k)
=
\frac{\partial f(x;\theta)}{\partial \theta}\Big|_{\theta=\theta_k}
\in \mathbb R^{1\times p}
\]
denote the parameter Jacobian at basin $k$. Under the local linear approximation, the reduction in within-basin target uncertainty is
\[
\Delta V_{\mathrm{intra}}(x)
=
\frac{1}{|\mathcal X_{\mathrm{tar}}|}
\sum_{k=1}^K
w_k\,
\frac{
\|J_k\Sigma_k J_x(\theta_k)^\top\|_2^2
}{
\sigma^2 + J_x(\theta_k)\Sigma_k J_x(\theta_k)^\top
}.
\]
This term is large when observing $y$ at $x$ is expected to substantially reduce predictive variance over the target region.

\paragraph{Inter-basin utility.}
The inter-basin gain is
\[
\Delta V_{\mathrm{inter}}(x)
=
V_{\mathrm{inter}}
-
\int
V_{\mathrm{inter}}^{+}(x,y)\,
p(y\mid x,\mathcal D_t)\,\mathrm dy,
\]
where $V_{\mathrm{inter}}^{+}(x,y)$ is the updated between-basin uncertainty after observing outcome $y$ at candidate $x$. The predictive distribution $p(y\mid x,\mathcal D_t)$ is the scalar mixture induced by the current basin approximation. Because the expectation is one-dimensional, we evaluate it efficiently using numerical quadrature.

\subsection{The Sequential Design Procedure}

Algorithm~\ref{alg:sequential_design} summarizes the full procedure. At each round, we first update the basin approximation from the current dataset, then score all remaining candidates using the target-aware acquisition function above, and finally select the highest-scoring affordable experiment.

\begin{algorithm}[t]
\caption{\method: Budget-aware sequential design for scaling-law fitting.}
\label{alg:sequential_design}
\begin{algorithmic}[1]
\REQUIRE Initial dataset $\mathcal D_0$, candidate pool $\mathcal X_{\mathrm{cand}}$, target region $\mathcal X_{\mathrm{tar}}$, cost function $c(\cdot)$, cost exponent $\alpha$
\FOR{$t = 0,1,2,\dots$ until budget is exhausted}
    \STATE $\{(\theta_k,\Sigma_k,w_k)\}_{k=1}^K \leftarrow \textsc{EstimateBasins}(\mathcal D_t)$
    \FOR{each $x \in \mathcal X_{\mathrm{cand}}$}
        \STATE $\Delta V_{\mathrm{intra}}(x) \leftarrow \textsc{IntraUtility}\!\left(x,\{(\theta_k,\Sigma_k,w_k)\}_{k=1}^K,\mathcal X_{\mathrm{tar}}\right)$
        \STATE $\Delta V_{\mathrm{inter}}(x) \leftarrow \textsc{InterUtility}\!\left(x,\{(\theta_k,\Sigma_k,w_k)\}_{k=1}^K,\mathcal X_{\mathrm{tar}}\right)$
        \STATE $S(x) \leftarrow \bigl(\Delta V_{\mathrm{intra}}(x)+\Delta V_{\mathrm{inter}}(x)\bigr)/c(x)^\alpha$
    \ENDFOR
    \STATE $x_{t+1} \leftarrow \arg\max_{x \in \mathcal X_{\mathrm{cand}}} S(x)$
    \STATE $y_{t+1} \leftarrow \textsc{RunExperiment}(x_{t+1})$
    \STATE $\mathcal D_{t+1} \leftarrow \mathcal D_t \cup \{(x_{t+1},y_{t+1})\}$
    \STATE $\mathcal X_{\mathrm{cand}} \leftarrow \mathcal X_{\mathrm{cand}} \setminus \{x_{t+1}\}$
\ENDFOR
\RETURN $\mathcal D_t$
\end{algorithmic}
\end{algorithm}

\section{Experiments}

\subsection{Experimental Setup}

\paragraph{Benchmark overview.}
We evaluate \method{} on a benchmark for budget-aware sequential design in scaling-law fitting, comprising 8 tasks and 65 scaling-law instances. Each instance specifies a parametric law family, a finite pool of runnable candidate experiments with associated costs, and a held-out target region for evaluation. The tasks cover diverse LLM scaling scenarios, including pre-training hyperparameter tuning, data allocation, architecture design, sparsity, and inference-time scaling. Table~\ref{tab:scalebench_tasks} summarizes the task-level statistics; detailed task descriptions and data sources are deferred to Appendix~\ref{app:task_details}.

\begin{table}[t]
\centering
% \small
\setlength{\tabcolsep}{6pt}
\renewcommand{\arraystretch}{1.05}
% \vspace{-7mm}
\begin{tabular}{lcccccc}
\toprule
\textbf{Task} & \textbf{\# Laws} & \textbf{Avg. Params} & \textbf{\# Train} & \textbf{\# Test} & \textbf{Target} & \textbf{Cost} \\
\midrule
\texttt{parallel} & 10 & 4.2 & 36 & 12 & $L(N, P)$ & $N$ \\
\texttt{vocab} & 10 & 6.8 & 1080 & 120 & $L(N, V, D)$ & $6ND$ \\
\texttt{domain} & 10 & 29.0 & 504 & 42 & $\{L_i(\mathrm{r})\}_{i=1}^5$ & $1$ \\
\texttt{moe} & 10 & 5.3 & 193 & 28 & $L(N, E)$ & $NE$ \\
\texttt{data\_con} & 10 & 7.0 & 161 & 21 & $L(N, D, U)$ & $6ND$ \\
\texttt{lr\&bsz} & 10 & 20.2 & 2702 & 117 & $L(l, b, N, D)$ & $6ND$ \\
\texttt{sparsity} & 4 & 5.0 & 70 & 18 & $L(P, N_2)$ & $6N_1D_1+6N_2D_2$ \\
\texttt{farseer} & 1 & 9.0 & 404 & 7 & $L(N, D)$ & $6ND$ \\
\bottomrule
\end{tabular}
\vspace{-1mm}
\caption{
\textbf{Benchmark task statistics.} Each task contains a collection of scaling-law instances for evaluating budget-aware sequential design on target-region extrapolation.
}
\label{tab:scalebench_tasks}
\end{table}

\paragraph{Law families and cost models.}
The benchmark spans a heterogeneous collection of scaling-law families, including classical power laws, log-space interaction models, compositional mixture laws, hyperparameter response surfaces, and several more expressive nonlinear forms; full parameterizations are deferred to Appendix~\ref{app:scalinglaw_recipe}. We assign each task a simple cost proxy aligned with its dominant resource. For dense-training settings (\texttt{data\_con}, \texttt{farseer}, \texttt{lr\&bsz}, and \texttt{vocab}), we use $6ND$. For the remaining tasks, we use task-specific proxies: $NE$ for \texttt{moe}, $N$ for \texttt{parallel}, $6N_1D_1 + 6N_2D_2$ for \texttt{sparsity}, and unit cost for \texttt{domain}, which varies mixture proportions rather than overall training scale.

\paragraph{Evaluation protocol and metrics.}
At the start of each episode, a method receives the candidate pool, target region, and candidate costs, but not the outcomes. It sequentially selects feasible unobserved candidates and refits the law after every observation using L-BFGS-B from 64 initializations. We repeat each method for 10 runs. We report target-region $R^2$ at $1\%$, $5\%$, and $10\%$ of total training cost for most tasks, and at $20\%$, $35\%$, and $50\%$ for \texttt{domain} and \texttt{sparsity}, whose costs are more uniform. Scores are aggregated over runs and instances within each task, clipped to $[-1,1]$, and reported as mean $\pm$ standard deviation. All methods use the same fitting procedure and differ only in experiment selection.

\paragraph{Baselines.}
We compare against five baselines. (1) \textbf{Random} uniformly samples the next experiment from the feasible unobserved candidates. (2) \textbf{Cheapest} always selects a minimum-cost feasible candidate, breaking ties uniformly at random. (3) \textbf{Cost Rand} samples each feasible unobserved candidate with probability proportional to $1/c(x)$. These three heuristics respectively capture uninformed exploration, an aggressive preference for cheap experiments, and a simple stochastic bias toward lower-cost candidates.

We further compare against two classical design criteria adapted to our nonlinear, cost-constrained setting. (4) \textbf{D-opt} selects the candidate that maximizes the increase in a D-optimality objective, which favors experiments that most reduce the overall volume of parameter uncertainty. In the locally linearized model, this corresponds to preferring points that most increase the log-determinant of the Fisher information matrix. (5) \textbf{V-opt} selects the candidate that maximizes a V-optimality objective, which favors experiments expected to most reduce predictive variance over the target region. In our implementation, both \textbf{D-opt} and \textbf{V-opt} locally linearize the nonlinear scaling law around the parameter estimate with the lowest MSE among the fitted solutions from the previous step, and differ only in the acquisition score computed from this linearization. To account for heterogeneous costs, each candidate score is normalized by $c(x)^\alpha$, with a single global $\alpha=0.4$ for all tasks and instances; Appendix~\ref{app:alpha_sensitivity} reports a sensitivity analysis.

For \textbf{D-opt}, \textbf{V-opt}, and \method, we use a short warm-start phase before the first criterion-based acquisition step: each method first selects the $2.5p$ lowest-cost candidates, where $p$ is the number of law parameters. The cost of these initialization points is counted toward the total budget, and Appendix~\ref{app:warm_start_analysis} reports both the warm-start cost fraction and an ablation over warm-start sizes. As a full-information reference, we also report \textbf{All Data}, obtained by fitting the same scaling law on the entire training set and evaluating the resulting target-region $R^2$.

\paragraph{Computational overhead.}
To quantitatively characterize the computational overhead of using \method{} in the sequential design loop, we profile the wall-clock time of each selection decision on \texttt{lr\&bsz}. This task has the largest candidate pool and relatively high-dimensional laws, making it one of the most computationally demanding settings in our benchmark. Table~\ref{tab:decision_time} decomposes the total decision time into refitting, basin construction, and acquisition scoring. The results indicate that a decision takes $1.260$ seconds on average, of which $1.136$ seconds (about $90\%$) is spent on refitting, while basin construction and acquisition scoring together account for only about $10\%$. Thus, the additional overhead specific to our acquisition procedure is small relative to the nonlinear refitting already required at each step. Appendix~\ref{app:computational_overhead} gives the scoring complexity.

\begin{table}[t]
\centering
\small
\setlength{\tabcolsep}{5pt}
\renewcommand{\arraystretch}{1.10}
\resizebox{\linewidth}{!}{%
\begin{tabular}{@{}lcccc@{}}
\toprule
\textbf{Component} & \textbf{Refitting} & \textbf{Basin construction} & \textbf{Acquisition scoring} & \textbf{Total decision time} \\
\midrule
Mean $\pm$ Std. (s) & $1.136 \pm 0.575$ & $0.065 \pm 0.055$ & $0.059 \pm 0.188$ & $1.260 \pm 0.582$ \\
\bottomrule
\end{tabular}
}
\caption{\textbf{Decision overhead} on \texttt{lr\&bsz}. Wall-clock decision time is reported separately from the executed training-run budget.}
\label{tab:decision_time}
\end{table}

\begin{figure}
    \centering
    % \vspace{-7mm}
    \includegraphics[width=\linewidth]{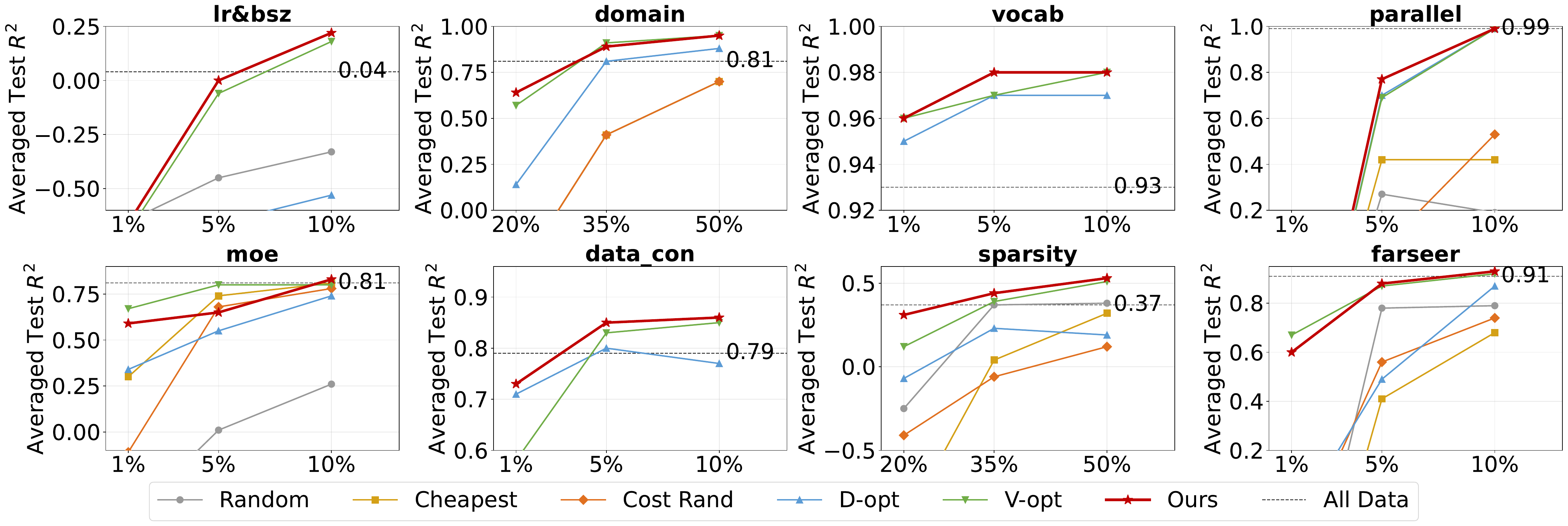}
    \vspace{-6mm}
    \caption{\textbf{Budget--accuracy curves} for target-region extrapolation. Mean target-region $R^2$ is shown as a function of consumed budget on the benchmark. \method{} reaches the best aggregate budget--accuracy trade-off and approaches the full-data reference using only a small fraction of the total experimental cost.}
    \label{fig:r2_vs_budget}
\end{figure}

\begin{table*}[t]
\centering
\small
\setlength{\tabcolsep}{2.0pt}
\renewcommand{\arraystretch}{1.10}
\begin{tabular}{@{}lcccccccc@{}}
\toprule
\textbf{Setting} & \textbf{lr\&bsz} & \textbf{domain} & \textbf{vocab} & \textbf{parallel} & \textbf{moe} & \textbf{data\_con} & \textbf{sparsity} & \textbf{farseer} \\
\midrule
1\% Random & \textbf{-0.65 {\tiny $\pm$ 0.49}} & -0.36 {\tiny $\pm$ 0.82} & 0.54 {\tiny $\pm$ 0.57} & -1.00 {\tiny $\pm$ 0.00} & -0.47 {\tiny $\pm$ 0.66} & -0.74 {\tiny $\pm$ 0.48} & -0.25 {\tiny $\pm$ 0.63} & -0.77 {\tiny $\pm$ 0.55} \\
1\% Cheapest & -0.92 {\tiny $\pm$ 0.29} & -0.36 {\tiny $\pm$ 0.82} & 0.30 {\tiny $\pm$ 0.71} & -1.00 {\tiny $\pm$ 0.00} & 0.30 {\tiny $\pm$ 0.57} & -0.58 {\tiny $\pm$ 0.64} & -0.95 {\tiny $\pm$ 0.14} & -0.89 {\tiny $\pm$ 0.17} \\
1\% Cost Rand & -0.89 {\tiny $\pm$ 0.31} & -0.36 {\tiny $\pm$ 0.82} & 0.86 {\tiny $\pm$ 0.38} & -1.00 {\tiny $\pm$ 0.00} & -0.11 {\tiny $\pm$ 0.70} & -0.38 {\tiny $\pm$ 0.70} & -0.41 {\tiny $\pm$ 0.52} & -0.25 {\tiny $\pm$ 0.63}\\
1\% D-opt  & -0.92 {\tiny $\pm$ 0.23} & 0.14 {\tiny $\pm$ 0.80} & 0.95 {\tiny $\pm$ 0.03} & -1.00 {\tiny $\pm$ 0.00} & 0.34 {\tiny $\pm$ 0.54} & 0.71 {\tiny $\pm$ 0.43} & -0.07 {\tiny $\pm$ 0.56} & -0.11 {\tiny $\pm$ 0.80} \\
1\% V-opt  & -0.70 {\tiny $\pm$ 0.47} & 0.57 {\tiny $\pm$ 0.66} & \textbf{0.96 {\tiny $\pm$ 0.02}} & -1.00 {\tiny $\pm$ 0.00} & \textbf{0.67 {\tiny $\pm$ 0.24}} & 0.58 {\tiny $\pm$ 0.53} & 0.12 {\tiny $\pm$ 0.53} & \textbf{0.67 {\tiny $\pm$ 0.22}} \\
1\% \method{} (ours)   & -0.66 {\tiny $\pm$ 0.53} & \textbf{0.64 {\tiny $\pm$ 0.58}} & \textbf{0.96 {\tiny $\pm$ 0.02}} & -1.00 {\tiny $\pm$ 0.00} & 0.59 {\tiny $\pm$ 0.39} & \textbf{0.73 {\tiny $\pm$ 0.37}} & \textbf{0.31 {\tiny $\pm$ 0.39}} & 0.60 {\tiny $\pm$ 0.11} \\
\midrule
5\% Random & -0.45 {\tiny $\pm$ 0.61} & 0.41 {\tiny $\pm$ 0.82} & 0.53 {\tiny $\pm$ 0.62} & 0.27 {\tiny $\pm$ 0.02} & 0.01 {\tiny $\pm$ 0.72} & -0.36 {\tiny $\pm$ 0.71} & 0.37 {\tiny $\pm$ 0.26} & 0.78 {\tiny $\pm$ 0.18} \\
5\% Cheapest & -0.88 {\tiny $\pm$ 0.34} & 0.41 {\tiny $\pm$ 0.82} & 0.53 {\tiny $\pm$ 0.62} & 0.42 {\tiny $\pm$ 0.03} & 0.74 {\tiny $\pm$ 0.19} & -0.56 {\tiny $\pm$ 0.65} & 0.04 {\tiny $\pm$ 0.46} & 0.41 {\tiny $\pm$ 0.40} \\
5\% Cost Rand & -0.79 {\tiny $\pm$ 0.43} & 0.41 {\tiny $\pm$ 0.82} & 0.89 {\tiny $\pm$ 0.27} & 0.03 {\tiny $\pm$ 0.87} & 0.68 {\tiny $\pm$ 0.33} & -0.39 {\tiny $\pm$ 0.65} & -0.06 {\tiny $\pm$ 0.52} & 0.56 {\tiny $\pm$ 0.38}\\
5\% D-opt  & -0.66 {\tiny $\pm$ 0.57} & 0.81 {\tiny $\pm$ 0.48} & 0.97 {\tiny $\pm$ 0.01    } & 0.70 {\tiny $\pm$ 0.52} & 0.55 {\tiny $\pm$ 0.34} & 0.80 {\tiny $\pm$ 0.17} & 0.23 {\tiny $\pm$ 0.32} & 0.49 {\tiny $\pm$ 0.18} \\
5\% V-opt  & -0.06 {\tiny $\pm$ 0.59} & \textbf{0.91 {\tiny $\pm$ 0.33}} & 0.97 {\tiny $\pm$ 0.00} & 0.69 {\tiny $\pm$ 0.53} & \textbf{0.80 {\tiny $\pm$ 0.05}} & 0.83 {\tiny $\pm$ 0.17} & 0.39 {\tiny $\pm$ 0.20} & 0.87 {\tiny $\pm$ 0.01} \\
5\% \method{} (ours)   & \textbf{0.00 {\tiny $\pm$ 0.59}} & 0.89 {\tiny $\pm$ 0.38} & \textbf{0.98 {\tiny $\pm$ 0.00}} & \textbf{0.77 {\tiny $\pm$ 0.39}} & 0.65 {\tiny $\pm$ 0.27} & \textbf{0.85 {\tiny $\pm$ 0.16}} & \textbf{0.44 {\tiny $\pm$ 0.17}} & \textbf{0.88 {\tiny $\pm$ 0.02}}\\
\midrule
10\% Random & -0.33 {\tiny $\pm$ 0.66} & 0.70 {\tiny $\pm$ 0.67} & 0.59 {\tiny $\pm$ 0.54} & 0.19 {\tiny $\pm$ 0.01} & 0.26 {\tiny $\pm$ 0.65} & 0.12 {\tiny $\pm$ 0.62} & 0.38 {\tiny $\pm$ 0.25} & 0.79 {\tiny $\pm$ 0.09} \\
10\% Cheapest & -0.80 {\tiny $\pm$ 0.46} & 0.70 {\tiny $\pm$ 0.67} & 0.55 {\tiny $\pm$ 0.63} & 0.42 {\tiny $\pm$ 0.03} & 0.81 {\tiny $\pm$ 0.14} & -0.40 {\tiny $\pm$ 0.64} & 0.32 {\tiny $\pm$ 0.20} & 0.68 {\tiny $\pm$ 0.18} \\
10\% Cost Rand & -0.79 {\tiny $\pm$ 0.39} & 0.70 {\tiny $\pm$ 0.67} & 0.83 {\tiny $\pm$ 0.34} & 0.53 {\tiny $\pm$ 0.81} & 0.78 {\tiny $\pm$ 0.23} & -0.22 {\tiny $\pm$ 0.67} & 0.12 {\tiny $\pm$ 0.38} & 0.74 {\tiny $\pm$ 0.11} \\
10\% D-opt  & -0.53 {\tiny $\pm$ 0.57} & 0.88 {\tiny $\pm$ 0.43} & 0.97 {\tiny $\pm$ 0.00} & \textbf{0.99 {\tiny $\pm$ 0.00}} & 0.74 {\tiny $\pm$ 0.14} & 0.77 {\tiny $\pm$ 0.16} & 0.19 {\tiny $\pm$ 0.41} & 0.87 {\tiny $\pm$ 0.03} \\
10\% V-opt  & 0.18 {\tiny $\pm$ 0.54} & \textbf{0.95 {\tiny $\pm$ 0.27}} & \textbf{0.98 {\tiny $\pm$ 0.00}} & \textbf{0.99 {\tiny $\pm$ 0.00}} & 0.80 {\tiny $\pm$ 0.11} & 0.85 {\tiny $\pm$ 0.13} & 0.51 {\tiny $\pm$ 0.15} & 0.92 {\tiny $\pm$ 0.01} \\
10\% \method{} (ours)   & \textbf{0.22 {\tiny $\pm$ 0.55}} & \textbf{0.95 {\tiny $\pm$ 0.28}} & \textbf{0.98 {\tiny $\pm$ 0.00}} & \textbf{0.99 {\tiny $\pm$ 0.00}} & \textbf{0.83 {\tiny $\pm$ 0.07}} & \textbf{0.86 {\tiny $\pm$ 0.11}} & \textbf{0.53 {\tiny $\pm$ 0.08}} & \textbf{0.93 {\tiny $\pm$ 0.00}} \\
\midrule 
All Data  & \textcolor{gray}{0.04 {\tiny $\pm$ 0.67}} & \textcolor{gray}{0.81 {\tiny $\pm$ 0.51}} & \textcolor{gray}{0.93 {\tiny $\pm$ 0.16}} & \textcolor{gray}{0.99 {\tiny $\pm$ 0.00}} & \textcolor{gray}{0.81 {\tiny $\pm$ 0.04}} & \textcolor{gray}{0.79 {\tiny $\pm$ 0.23}} & \textcolor{gray}{0.37 {\tiny $\pm$ 0.10}} & \textcolor{gray}{0.91 {\tiny $\pm$ 0.01}} \\
\bottomrule
\end{tabular}
% \vspace{-2mm}
\caption{
\textbf{Task-level target-region performance.} Each cell reports the mean and standard deviation of target-region $R^2$ aggregated over all scaling-law instances within the corresponding task. Higher is better, and $R^2$ is clipped to $[-1,1]$. Budgets are $1\%$, $5\%$, $10\%$ for most tasks, and $20\%$, $35\%$, $50\%$ for \texttt{domain} and \texttt{sparsity}.
}
\label{tab:task_breakdown}
\end{table*}

\subsection{Results Analysis}

\paragraph{Overall trends.}
Figure~\ref{fig:r2_vs_budget} and Table~\ref{tab:task_breakdown} show that \method{} delivers the best aggregate performance. Its advantage is most pronounced in the low-budget regime, where experiment selection matters most: at $1\%$ budget, it performs best on \texttt{domain}, \texttt{data\_con}, and \texttt{sparsity}, matches the top result on \texttt{vocab}, and remains competitive on \texttt{moe} and \texttt{farseer}. As the budget increases, the advantage becomes more consistent. At $5\%$ budget, it achieves the best result on six of the eight tasks, and at $10\%$ budget, it matches or outperforms all baselines on every task.

A consistent pattern is that model-aware design substantially outperforms simple budget heuristics. Across most tasks and budgets, \textbf{Random}, \textbf{Cheapest}, and \textbf{Cost Rand} lag far behind, especially under tight budgets, showing that neither uninformed exploration nor naive cost preference is sufficient for reliable target-region extrapolation.

\begin{figure*}[t]
    \centering
    \includegraphics[width=0.99\textwidth]{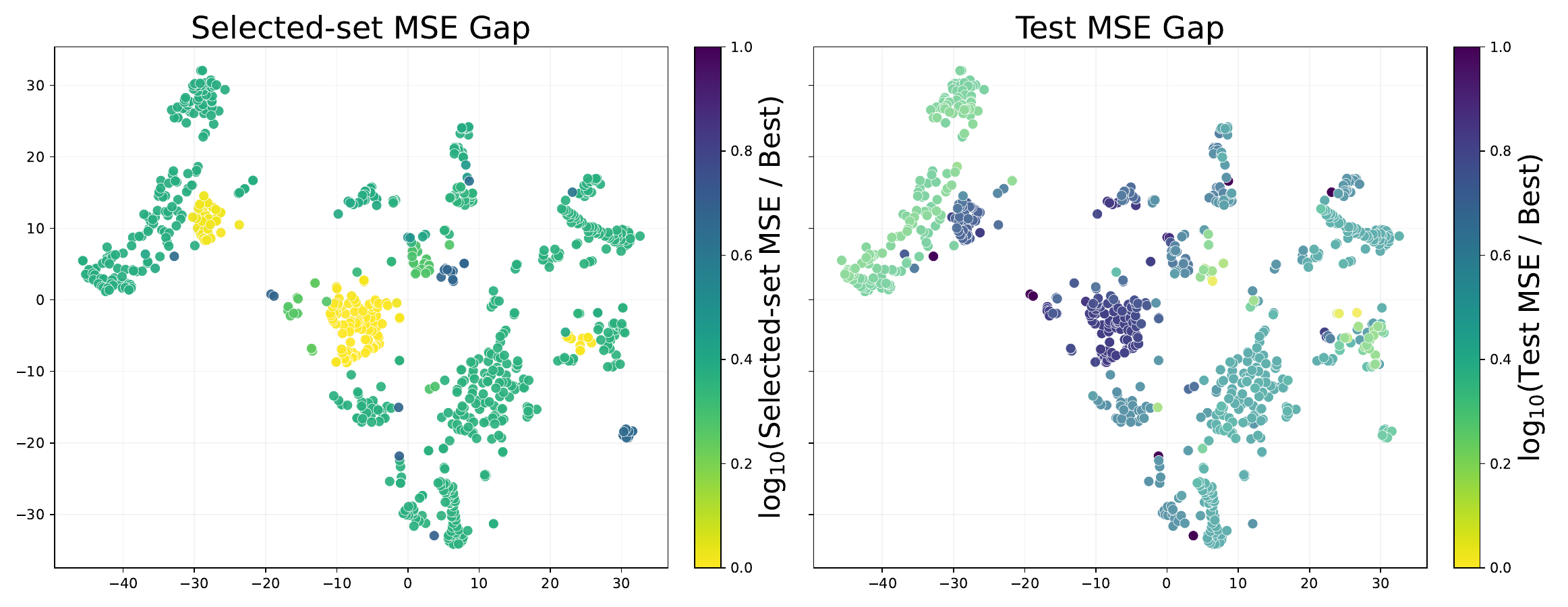}
    \vspace{-2mm}
    \caption{
    \textbf{Local optima and extrapolation mismatch} in \texttt{lr\&bsz}. Parameter-space visualization for one \texttt{lr\&bsz} scaling law (\texttt{sl\_5}) after fitting on the cheapest $12\%$ of training points from $2048$ initializations.
    We embed the fitted parameters with t-SNE and color each solution by its MSE on the selected points (left) or on the held-out test region (right).
    Multiple separated clusters indicate many local optima, while the mismatch between the two colorings shows that low error on the observed low-cost points does not reliably imply low error on the high-cost extrapolation region. \method{} achieves $0.71$ averaged test $R^2$ within $5\%$ budget compared with $0.57$ for V-opt and $0.16$ for D-opt.
    }
    \label{fig:lrbsz_tsne}
\end{figure*}

\paragraph{Comparison to design-based baselines.}
Among the design-based baselines, \textbf{D-opt} is already much stronger than the simple heuristics, highlighting the value of exploiting local sensitivity information. \textbf{V-opt} is closer in spirit to \method: both are prediction-oriented, but \textbf{V-opt} relies on a single local linearization around the current best-fit parameter, whereas \method{} maintains a mixture-of-Gaussians representation over multiple plausible parameter regions. Empirically, our method is generally more robust and achieves better overall performance, especially at moderate and higher budgets. This gap is most pronounced when the fitting landscape contains multiple plausible basins and the target region is strongly extrapolative, so that the current best fit need not be the right expansion point for design. Figure~\ref{fig:lrbsz_tsne} illustrates such a case on one \texttt{lr\&bsz} scaling law using t-SNE visualization \citep{van2008visualizing}.

\paragraph{Task and budget dependence.}
Performance varies substantially across tasks, reflecting the heterogeneous structure of the benchmark. Because each entry in Table~\ref{tab:task_breakdown} averages over multiple scaling-law instances within a task, these results should be interpreted as robustness over a heterogeneous family of fitting problems rather than performance on any single law. Some tasks are relatively easy on average: on \texttt{vocab}, all design-based methods already perform strongly at low budget, and the gap nearly closes by $5\%$ budget; similarly, \texttt{parallel} is close to saturated once the budget becomes moderate. By contrast, \texttt{lr\&bsz}, \texttt{domain}, \texttt{data\_con}, and \texttt{sparsity} remain more challenging on average and show larger separation among methods across a wider range of budgets. Overall, low-budget performance is the main differentiator across methods, while higher-budget results reveal which approaches remain robust on the harder and more heterogeneous tasks.

\paragraph{Why can some methods outperform \textbf{All Data}?}
Some budgeted methods occasionally outperform \textbf{All Data}. This is not contradictory, because \textbf{All Data} is not an oracle upper bound: it is simply the test-region $R^2$ obtained by fitting the same parametric law on the full training set. Since the target region is fully extrapolative and concentrated in the high-cost regime, the best fit on all observed training points need not be the best fit for the target region. Under model misspecification, adding more training points can even hurt extrapolation if those points are concentrated in regions whose trends are less aligned with the high-cost behavior of interest.

To test whether this gain merely comes from avoiding cheap runs, we compare \method{} with fits using only the cheapest or most expensive half of each candidate pool. Table~\ref{tab:cheap_expensive_diagnostic} shows that \method{} outperforms both diagnostics on average and beats the expensive-half fit on six of eight tasks. The gain therefore comes from target-aware selection, not a simple preference for expensive runs. More broadly, when the goal is to predict much larger configurations, blindly fitting all available pilot runs may be suboptimal.

\begin{table}[t]
\centering
\small
\setlength{\tabcolsep}{3pt}
\renewcommand{\arraystretch}{1.10}
\resizebox{\linewidth}{!}{%
\begin{tabular}{@{}lccccccccc@{}}
\toprule
\textbf{Metric} & \textbf{lr\&bsz} & \textbf{domain} & \textbf{vocab} & \textbf{parallel} & \textbf{moe} & \textbf{data\_con} & \textbf{sparsity} & \textbf{farseer} & \textbf{Average} \\
\midrule
\method{} @ final checkpoint & $0.22$ & $0.95$ & $0.98$ & $0.99$ & $0.83$ & $0.86$ & $0.53$ & $0.93$ & $0.79$ \\
Cheap half & $-0.89$ & $0.79$ & $0.84$ & $1.00$ & $0.42$ & $0.54$ & $-0.67$ & $0.51$ & $0.32$ \\
Expensive half & $0.24$ & $0.79$ & $0.42$ & $1.00$ & $0.33$ & $0.84$ & $0.48$ & $0.79$ & $0.61$ \\
\bottomrule
\end{tabular}
}
\vspace{-1mm}
\caption{\textbf{Cheap and expensive half diagnostic.} Target-region $R^2$ for the final-checkpoint design selected by \method{} versus fits using only the cheapest or most expensive half of the candidate pool. Higher is better.}
\label{tab:cheap_expensive_diagnostic}
\end{table}

\subsection{Ablation Study}

\begin{table*}[t]
\centering
% \vspace{-3mm}
\small
\setlength{\tabcolsep}{2.0pt}
\renewcommand{\arraystretch}{1.10}
\begin{tabular}{@{}lcccccccc@{}}
\toprule
\textbf{Setting} & \textbf{lr\&bsz} & \textbf{domain} & \textbf{vocab} & \textbf{parallel} & \textbf{moe} & \textbf{data\_con} & \textbf{sparsity} & \textbf{farseer} \\
\midrule
1\% w/o $\Delta V_{\mathrm{inter}}$ & -0.70 {\tiny $\pm$ 0.48} & 0.64 {\tiny $\pm$ 0.58} & 0.96 {\tiny $\pm$ 0.02} & -1.00 {\tiny $\pm$ 0.00} & 0.55 {\tiny $\pm$ 0.44} & 0.69 {\tiny $\pm$ 0.39} & 0.08 {\tiny $\pm$ 0.55} & 0.60 {\tiny $\pm$ 0.11} \\
1\% w/o $\Delta V_{\mathrm{intra}}$ & -0.78 {\tiny $\pm$ 0.44} & 0.39 {\tiny $\pm$ 0.76} & 0.96 {\tiny $\pm$ 0.02} & -1.00 {\tiny $\pm$ 0.00} & 0.43 {\tiny $\pm$ 0.63}  & -0.01 {\tiny $\pm$ 0.79} & -0.07 {\tiny $\pm$ 0.71} & -0.41 {\tiny $\pm$ 0.73} \\
1\% \method{} (ours) & -0.66 {\tiny $\pm$ 0.53} & 0.64 {\tiny $\pm$ 0.58} & 0.96 {\tiny $\pm$ 0.02} & -1.00 {\tiny $\pm$ 0.00} & 0.59 {\tiny $\pm$ 0.39} & 0.73 {\tiny $\pm$ 0.37} & 0.31 {\tiny $\pm$ 0.39} & 0.60 {\tiny $\pm$ 0.11} \\
\midrule
5\% w/o $\Delta V_{\mathrm{inter}}$ & -0.02 {\tiny $\pm$ 0.60} & 0.89 {\tiny $\pm$ 0.38} & 0.97 {\tiny $\pm$ 0.00} & 0.77 {\tiny $\pm$ 0.39} & 0.67 {\tiny $\pm$ 0.23} & 0.83 {\tiny $\pm$ 0.17} & 0.33 {\tiny $\pm$ 0.30} & 0.88 {\tiny $\pm$ 0.02} \\
5\% w/o $\Delta V_{\mathrm{intra}}$ & -0.51 {\tiny $\pm$ 0.53} & 0.86 {\tiny $\pm$ 0.43} & 0.98 {\tiny $\pm$ 0.00} & 0.07 {\tiny $\pm$ 0.90} & 0.34 {\tiny $\pm$ 0.40} & 0.65 {\tiny $\pm$ 0.47} & 0.37 {\tiny $\pm$ 0.39} & 0.84 {\tiny $\pm$ 0.09} \\
5\% \method{} (ours) & 0.00 {\tiny $\pm$ 0.59} & 0.89 {\tiny $\pm$ 0.38} & 0.98 {\tiny $\pm$ 0.00} & 0.77 {\tiny $\pm$ 0.39} & 0.65 {\tiny $\pm$ 0.27} & 0.85 {\tiny $\pm$ 0.16} & 0.44 {\tiny $\pm$ 0.17} & 0.88 {\tiny $\pm$ 0.02} \\
\midrule
10\% w/o $\Delta V_{\mathrm{inter}}$ & 0.20 {\tiny $\pm$ 0.55} & 0.95 {\tiny $\pm$ 0.28} & 0.98 {\tiny $\pm$ 0.00} & 0.99 {\tiny $\pm$ 0.01} & 0.83 {\tiny $\pm$ 0.08} & 0.85 {\tiny $\pm$ 0.12} & 0.51 {\tiny $\pm$ 0.13} & 0.93 {\tiny $\pm$ 0.00} \\
10\% w/o $\Delta V_{\mathrm{intra}}$ & -0.13 {\tiny $\pm$ 0.60} & 0.91 {\tiny $\pm$ 0.35} & 0.98 {\tiny $\pm$ 0.00} & 0.99 {\tiny $\pm$ 0.00} & 0.71 {\tiny $\pm$ 0.22} & 0.80 {\tiny $\pm$ 0.25} & 0.48 {\tiny $\pm$ 0.19} & 0.89 {\tiny $\pm$ 0.04} \\
10\% \method{} (ours) & 0.22 {\tiny $\pm$ 0.55} & 0.95 {\tiny $\pm$ 0.28} & 0.98 {\tiny $\pm$ 0.00} & 0.99 {\tiny $\pm$ 0.00} & 0.83 {\tiny $\pm$ 0.07} & 0.86 {\tiny $\pm$ 0.11} & 0.53 {\tiny $\pm$ 0.08} & 0.93 {\tiny $\pm$ 0.00} \\
\bottomrule
\end{tabular}
% \vspace{-3mm}
\caption{
\textbf{Acquisition-function ablation.} We remove either $\Delta V_{\mathrm{inter}}$ or $\Delta V_{\mathrm{intra}}$ from the acquisition score and report the target-region $R^2$ at different budget levels. Each cell shows the mean and standard deviation aggregated over all scaling-law instances. Higher is better. Budgets are $1\%$, $5\%$, $10\%$ for most tasks, and $20\%$, $35\%$, $50\%$ for \texttt{domain} and \texttt{sparsity}.
}
\label{tab:ablation_breakdown}
\end{table*}

To understand which parts of our acquisition function are responsible for the observed gains, we perform an ablation study by removing the two terms in the MSPE decomposition, $\Delta V_{\mathrm{inter}}$ and $\Delta V_{\mathrm{intra}}$. Here, $V_{\mathrm{inter}}$ measures uncertainty induced by disagreement across different basins, while $V_{\mathrm{intra}}$ measures uncertainty within each basin due to local parameter variation around a mode. Ablating them separately allows us to isolate the contributions of cross-basin and within-basin uncertainty to sequential experiment selection.

\paragraph{Ablation results.}
Table~\ref{tab:ablation_breakdown} shows that both terms contribute, but not equally. Removing $\Delta V_{\mathrm{intra}}$ causes the larger and more consistent degradation, especially on \texttt{data\_con}, \texttt{farseer}, \texttt{moe}, and \texttt{lr\&bsz}, indicating that within-basin uncertainty is the dominant signal for effective sequential design. By contrast, removing $\Delta V_{\mathrm{inter}}$ usually leads to a smaller drop and in some cases leaves performance nearly unchanged, suggesting that cross-basin disagreement is more task-dependent.

This difference is consistent with the roles of the two terms. $\Delta V_{\mathrm{intra}}$ remains useful even after the method has identified a plausible parameter region, because it continues to refine uncertainty within that basin. $\Delta V_{\mathrm{inter}}$ is most helpful when several basins remain plausible and induce different target-region predictions, which occurs more often under tighter budgets and on more heterogeneous tasks such as \texttt{sparsity} and \texttt{data\_con}. Overall, the full method is the most robust across tasks and budget levels, indicating that the two terms are complementary: $\Delta V_{\mathrm{intra}}$ provides the dominant signal, while $\Delta V_{\mathrm{inter}}$ yields additional gains when basin ambiguity is substantial.

\section{Conclusions}

Our work formulates scaling-law fitting as a budget-aware sequential experimental design problem, where each candidate run incurs a cost and the objective is to maximize predictive accuracy in a high-cost target region. We propose \method{}, an uncertainty-aware acquisition strategy that selects experiments according to their value for target-region extrapolation. Across a diverse benchmark of scaling-law tasks, \method{} generally outperforms random, heuristic, and classical design-based baselines, and often approaches full-data performance using only a small fraction of the original training budget. These results demonstrate that treating scaling-law fitting as a problem of experimental design and budget allocation leads to new algorithmic insights.

We also note that \method{} may be less reliable under severe model misspecification, particularly when abrupt phase transitions are neither visible in the pilot region nor represented by the chosen law family. Moreover, its one-step acquisition rule does not explicitly optimize long-horizon budget allocation. More robust uncertainty approximations and batched or multi-step budget-aware design are promising directions for future work.

\clearpage

\bibliography{reference}
\bibliographystyle{colm2026_conference}
\newpage
\appendix
\section{Use of LLMs}

We employ large language models (LLMs) exclusively for the purpose of assisting in the drafting and refinement of our manuscripts, with the objective of enhancing clarity and coherence.
\section{Detailed Statistics of the Benchmark}

\subsection{Task Collection Details}
\label{app:task_details}

Our experimental benchmark covers a diverse set of practical LLM scaling scenarios, including pre-training hyperparameter tuning, data allocation, architecture design, and inference-time scaling. The first six tasks are drawn from SLDBench \citep{lin2026can}, and the remaining two from the original papers; we also refer to surveys and related literature for task coverage and context \citep{sengupta2025how}. For the six SLDBench datasets, we use the top-10 law forms on the leaderboard ranked by mean $R^2$. We exclude \textit{Supervised Finetuning Scaling Law} due to its very limited data and \textit{U-shaped Scaling Law} due to severe misspecification that yields poor fits even on the full training set.

The collected tasks are: (1) \textit{Parallel Scaling Law} (\texttt{parallel}), which studies the effect of parallelism $P$ and model size $N$ on language modeling loss; this setting creates $P$ augmentations of an input and aggregates their outputs, conceptually similar to Best-of-$N$ \citep{chen2025parallelscalinglawlanguage}. (2) \textit{Vocabulary Scaling Law} (\texttt{vocab}), which models unigram-normalized loss as a function of non-vocabulary model size $N$, vocabulary size $V$, and dataset size $D$ \citep{tao2024scaling}. (3) \textit{Domain Mixture Scaling Law} (\texttt{domain}), which models domain-specific pre-training loss as a function of the mixture proportions of training domains \citep{ye2025data}. (4) \textit{Mixture of Experts Scaling Law} (\texttt{moe}), which relates loss to the number of dense parameters $N$ and experts $E$ \citep{krajewski2024scalinglawsfinegrainedmixture}. (5) \textit{Data Constrained Scaling Law} (\texttt{data\_con}), which models pre-training loss using network size $N$, dataset size $D$, and the number of unique tokens $U$. (6) \textit{Learning Rate and Batch Size Scaling Law} (\texttt{lr\&bsz}), adapted from the Step Law \citep{li2025predictable}, which models pre-training loss as a function of learning rate $l$, batch size $b$, dataset size $D$, and network size $N$. (7) \textit{Sparsity Scaling Law} (\texttt{sparsity}), which models test loss based on the ratio $P$ between total model size $N_1$ and active parameters $N_2$ \citep{liew2025scaling}. (8) \textit{Farseer Scaling Law} (\texttt{farseer}), which extends the Chinchilla-style formulation \citep{hoffmann2022trainingcomputeoptimallargelanguage} to predict loss from model size $N$ and training data $D$ \citep{li2025predictablescaleiifarseer}.

\subsection{Parametric Forms of Collected Scaling Laws}
\label{app:scalinglaw_recipe}

This appendix summarizes the parametric scaling-law families used in our benchmark. Our goal is not to advocate a single canonical law form, but to evaluate budget-aware sequential design across a diverse set of realistic fitting regimes. The collected laws therefore span classical power-law formulations, log-space interaction models, compositional mixture laws, hyperparameter response surfaces, and several more expressive nonlinear forms drawn from prior scaling-law studies. For the six tasks inherited from SLDBench, we use the top-ranked law families on the public leaderboard by mean $R^2$; for the remaining tasks, we adopt the parametric forms proposed in the corresponding original papers. Together, these forms define the nonlinear fitting landscapes on which all acquisition methods are evaluated.

\begin{longtable}{>{\raggedright\arraybackslash}p{0.08\textwidth} >{\raggedright\arraybackslash}p{0.58\textwidth} >{\centering\arraybackslash}p{0.12\textwidth} >{\centering\arraybackslash}p{0.12\textwidth}}
\caption{\textbf{Collected scaling laws.} Scaling laws grouped by task.}\label{tab:scaling-laws}\\
\toprule
ID & Parametric Form  & \# Params & \textbf{All Data} test $R^2$ \\
\midrule
\endfirsthead

\toprule
ID & Parametric Form & \# Params & \textbf{All Data} test $R^2$ \\
\midrule
\endhead

\bottomrule
\endfoot

\multicolumn{4}{l}{\textbf{\texttt{data\_con}}} \\
\midrule
\texttt{sl\_1} & $L(N,D,U)=\frac{A}{N^\alpha}+\frac{B}{D^\beta}+E\,U^\gamma N^\delta$ & 7 & 0.94 {\tiny $\pm$ 0.00} \\
\texttt{sl\_2} & $L(N,D,U)=a+bU^p+cN^q+dD^r$ & 7 & 0.73 {\tiny $\pm$ 0.11} \\
\texttt{sl\_3} & $L(N,D,U)=\frac{A}{N_{\mathrm{eff}}^\alpha}+\frac{B}{D_{\mathrm{eff}}^\alpha}+C,\quad U_N=\min(\rho U,N),\ R_N=\max(N/U_N-1,0),\ N_{\mathrm{eff}}=U_N+\tau_NU_N(1-e^{-R_N/\tau_N}),\ D_{\mathrm{eff}}=U+\tau_DU(1-e^{-(D/U-1)/\tau_D})$ & 7 & 0.89 {\tiny $\pm$ 0.00} \\
\texttt{sl\_4} & $L(N,D,U)=L_0+A M_n^{-a}+B T_{\mathrm{eff},n}^{-b},\quad q=\frac{T_n}{sU_nM_n^d},\ T_{\mathrm{eff},n}=\frac{T_n}{1+q}$ & 7 & 0.95 {\tiny $\pm$ 0.01} \\
\texttt{sl\_5} & $L(N,D,U)=\frac{A}{N^\alpha}+\frac{B}{D_{\mathrm{eff}}^\beta}+E,\quad D_{\mathrm{eff}}=U^\gamma D^{1-\gamma}$ & 6 & 0.84 {\tiny $\pm$ 0.00} \\
\texttt{sl\_6} & $L(N,D,U)=E+A N^{-\alpha}+B D_{\mathrm{eff}}^{-\beta},\quad D_{\mathrm{eff}}=\frac{D}{1+C\max(D/U-1,0)^cN^d}$ & 8 & 0.88 {\tiny $\pm$ 0.08} \\
\texttt{sl\_7} & $L(N,D,U)=L_0+A(NU)^{\alpha_{pu}}+BD^{\alpha_t}+CN^{\alpha_p}$ & 7 & 0.94 {\tiny $\pm$ 0.03} \\
\texttt{sl\_8} & $L(N,D,U)=a+\frac{b}{D^\alpha}+\frac{c}{N^\beta}+d\left|\log\!\left(\frac{U}{D}+1\right)\right|^\gamma$ & 7 & -0.07 {\tiny $\pm$ 0.19} \\
\texttt{sl\_9} & $L(N,D,U)=\frac{A}{N^\alpha}+\frac{B}{D^\beta}\left(1+\frac{C}{U^\gamma}\right)+L_{\inf}$ & 7 & 0.84 {\tiny $\pm$ 0.08} \\
\texttt{sl\_10} & $L(N,D,U)=L_0+A N^{-a}+B\left(D^{-bq}+(kU)^{-bq}\right)^{1/q}$ & 7 & 0.81 {\tiny $\pm$ 0.03} \\
\addlinespace
\midrule
\multicolumn{4}{l}{\textbf{\texttt{domain}}} \\
\midrule
\texttt{sl\_1} & $L_i(\mathrm r)=a_i+b_i\log(r_i+\varepsilon)+\sum_{j\neq i}c_{ij}r_j$ & 30 & 1.00 {\tiny $\pm$ 0.00} \\
\texttt{sl\_2} & $L_i(\mathrm r)=A_i(r_i+\varepsilon_i)^{-\alpha_i}\exp\!\left(\sum_{j\neq i}w_{ij}r_j\right)$ & 35 & 1.00 {\tiny $\pm$ 0.00} \\
\texttt{sl\_3} & $L_i(\mathrm r)=\mathrm{base}_i+\mathrm{coeff}_i\,r_i^{\mathrm{exp}_i}+\sum_{j\neq i}W_{ij}r_j$ & 35 & 1.00 {\tiny $\pm$ 0.00} \\
\texttt{sl\_4} & $L_i(\mathrm r)=\exp\!\left(\sum_k C_{ik}r_k^{\alpha_k}+\mathrm{bias}_i\right)$ & 35 & 1.00 {\tiny $\pm$ 0.00} \\
\texttt{sl\_5} & $L_i(\mathrm r)=b_i+\sum_j W_{ij}r_j^{\alpha_j}$ & 35 & 1.00 {\tiny $\pm$ 0.00} \\
\texttt{sl\_6} & $L_i(\mathrm r)=C_i+A_i\left(\sum_j T_{ij}r_j\right)^{-\alpha_i}$ & 35 & -0.81 {\tiny $\pm$ 0.58} \\
\texttt{sl\_7} & $L_i(\mathrm r)=\mathrm{intercept}_i+\sum_j\left(c^{\mathrm{lin}}_{ij}r_j+c^{\log}_{ij}\log(r_j+\varepsilon)\right)$ & 40 & 1.00 {\tiny $\pm$ 0.00} \\
\texttt{sl\_8} & $L_i(\mathrm r)=c_i-a_i r_i^{b_i}$ & 15 & 1.00 {\tiny $\pm$ 0.00} \\
\texttt{sl\_9} & $L_i(\mathrm r)=a_i+b_i\log(r_i+\varepsilon)+c_i\bigl[\log(r_i+\varepsilon)\bigr]^2$ & 15 & 1.00 {\tiny $\pm$ 0.00} \\
\texttt{sl\_10} & $L_i(\mathrm r)=a_i+\frac{b_i}{r_i+\varepsilon_i}$ & 15 & 1.00 {\tiny $\pm$ 0.00} \\
\addlinespace
\midrule
\multicolumn{4}{l}{\textbf{\texttt{farseer}}} \\
\midrule
\texttt{sl\_1} & $L(N,D)=e^{sN^q+S}+e^{BN^b+Q}\,D^{-e^{AN^a+E}}$ & 9 & 0.92 {\tiny $\pm$ 0.01} \\
\addlinespace
\midrule
\multicolumn{4}{l}{\textbf{\texttt{lr\&bsz}}} \\
\midrule
\texttt{sl\_1} & $L(l,b,N,D)=\exp\!\bigl(\operatorname{poly}_2(\log l,\log b,\log D,\log N)\bigr)$ & 15 & 0.61 {\tiny $\pm$ 0.19} \\
\texttt{sl\_2} & $L(l,b,N,D)=L_{\inf}+C_p e^{-a_pn}+C_d e^{-a_ds}+C_{dp}e^{-a_{dp}(s-kn)}+C_{bb}e^{-a_{bb}v}+c_L\Delta u^2+c_B\Delta v^2+2\rho\sqrt{c_Lc_B}\,\Delta u\,\Delta v,\quad n=\log N,\ s=\log D,\ u=\log l,\ v=\log b$ & 26 & -0.15 {\tiny $\pm$ 0.30} \\
\texttt{sl\_3} & $L(l,b,N,D)=E+AN^{-\alpha}+BD^{-\beta}+\frac{F}{N^{w_N}D^{w_D}}+C_{\mathrm{eff}}(\log l-\mathrm{opt}_l)^2+G_{\mathrm{eff}}(\log b-\mathrm{opt}_b)^2$ & 24 & -0.99 {\tiny $\pm$ 0.04} \\
\texttt{sl\_4} & $L(l,b,N,D)=\exp\!\big(w_0+w_1\log l+w_2\log b+w_3\log D+w_4\log N+w_5\log^2l+w_6\log^2b+w_7\log^2D+w_8\log^2N+w_9\log l\log b+w_{10}\log l\log D+w_{11}\log l\log N+w_{12}\log b\log D+w_{13}\log b\log N+w_{14}\log D\log N+w_{15}(\log D-\log N)+w_{16}/b+w_{17}/b^2+w_{18}/D+w_{19}/N\big)$ & 20 & -0.88 {\tiny $\pm$ 0.36} \\
\texttt{sl\_5} & $L(l,b,N,D)=L_0+A_Ne^{-a_Nn}+A_De^{-a_Ds}+A_Be^{-a_Bv}+A_Re^{-a_R(s-n)^2}+A_Xe^{-a_X(s-v)}+c_{\mathrm{lr},0}e^{-w_bv-w_nn-w_ss}(u-u_\star)^2,\quad u=\log l,\ v=\log b,\ s=\log D,\ n=\log N$ & 19 & -0.10 {\tiny $\pm$ 0.66} \\
\texttt{sl\_6} & $L(l,b,N,D)=L_{\inf}+\exp\!\big(w_0+w_d\log D+w_p\log N+w_{dp}\log D\log N+w_{\mathrm{lr}}\log l+w_{\mathrm{lr}^2}\log^2l+w_{\mathrm{bsz}}\log b+w_{\mathrm{bsz}^2}\log^2b+w_{\mathrm{lr,bsz}}\log l\log b+w_{\mathrm{lr},D}\log l\log D+w_{\mathrm{lr},N}\log l\log N+w_{\mathrm{bsz},D}\log b\log D+w_{\mathrm{bsz},N}\log b\log N\big)$ & 14 & 0.47 {\tiny $\pm$ 0.14} \\
\texttt{sl\_7} & $L(l,b,N,D)=E+\exp\!\big(w_{1,0}+w_{1,1}x_1+w_{1,2}x_2+w_{1,3}x_3+w_{1,4}x_4+w_{1,5}x_1^2+w_{1,6}x_2^2+w_{1,7}x_3^2+w_{1,8}x_4^2+w_{1,9}x_1x_2+w_{1,10}x_1x_3+w_{1,11}x_1x_4+w_{1,12}x_2x_3+w_{1,13}x_2x_4+w_{1,14}x_3x_4\big)+\exp\!\big(w_{2,0}+w_{2,1}x_1+w_{2,2}x_2+w_{2,3}x_3+w_{2,4}x_4+w_{2,5}x_1^2+w_{2,6}x_2^2+w_{2,7}x_3^2+w_{2,8}x_4^2+w_{2,9}x_1x_2+w_{2,10}x_1x_3+w_{2,11}x_1x_4+w_{2,12}x_2x_3+w_{2,13}x_2x_4+w_{2,14}x_3x_4\big),\ x_1=\log l,\ x_2=\log b,\ x_3=\log D,\ x_4=\log N$ & 31 & 0.75 {\tiny $\pm$ 0.19} \\
\texttt{sl\_8} & $L(l,b,N,D)=L_0+c_Pe^{-a_Pn}+c_De^{-a_Ds}+c_Re^{-a_R(s-n)}+k_{\mathrm{lr}}\delta_{\mathrm{lr}}^2(1+a_{\mathrm{lr}}\tanh\delta_{\mathrm{lr}})+k_{\mathrm{ns}}\delta_{\mathrm{ns}}^2(1+a_{\mathrm{ns}}\tanh\delta_{\mathrm{ns}})+k_{\mathrm{dp}}((s-n)-\delta_0)^2,\quad n=\log N,\ s=\log D$ & 20 & 0.49 {\tiny $\pm$ 0.33} \\
\texttt{sl\_9} & $L(l,b,N,D)=\operatorname{poly}_2(\log_{10}l,\log_{10}b,\log_{10}D,\log_{10}N)$ & 15 & 0.36 {\tiny $\pm$ 0.09} \\
\texttt{sl\_10} & $L(l,b,N,D)=\operatorname{poly}_2(\log l,\log b,\log D,\log N)+w_DD^{-1/2}+w_NN^{-1/2}+w_bb^{-1}$ & 18 & -0.07 {\tiny $\pm$ 0.64} \\
\addlinespace
\midrule
\multicolumn{4}{l}{\textbf{\texttt{moe}}} \\
\midrule
\texttt{sl\_1} & $L(N,E)=L_{\inf}+\frac{B}{N^\alpha E^\beta}$ & 4 & 0.83 {\tiny $\pm$ 0.00} \\
\texttt{sl\_2} & $L(N,E)=L+K(N^\alpha E^\beta)^{-\gamma}$ & 5 & 0.83 {\tiny $\pm$ 0.00} \\
\texttt{sl\_3} & $L(N,E)=\frac{A N^\alpha}{1+BE^\beta}+C N^{0.6\alpha}+D$ & 6 & 0.90 {\tiny $\pm$ 0.00} \\
\texttt{sl\_4} & $L(N,E)=\frac{a}{N^\alpha(1+bE)^\gamma}+c+d\bigl(\log N-0.4\log(1+E)\bigr)$ & 6 & 0.83 {\tiny $\pm$ 0.00} \\
\texttt{sl\_5} & $L(N,E)=p_0+\exp\!\bigl(p_1+p_2\log E+p_3\log N+p_4\log E\log N\bigr)+p_5\log E$ & 6 & 0.80 {\tiny $\pm$ 0.00} \\
\texttt{sl\_6} & $L(N,E)=aN^{-b}(1+cE^{-d})+e+\frac{f}{EN^{0.05}}$ & 6 & 0.83 {\tiny $\pm$ 0.00} \\
\texttt{sl\_7} & $L(N,E)=p_0E^{p_1}N^{p_2}+p_3N^{p_4}+p_5$ & 6 & 0.74 {\tiny $\pm$ 0.00} \\
\texttt{sl\_8} & $L(N,E)=aN^bE^c+d$ & 4 & 0.83 {\tiny $\pm$ 0.00} \\
\texttt{sl\_9} & $L(N,E)=c_0+A(NE^g)^{-a}$ & 4 & 0.83 {\tiny $\pm$ 0.00} \\
\texttt{sl\_10} & $L(N,E)=\mathrm{bias}+A(N/10^9)^{-\alpha}\left(\frac{1+BE^\gamma}{1+B}\right)^{-\beta}$ & 6 & 0.77 {\tiny $\pm$ 0.00} \\
\addlinespace
\midrule
\multicolumn{4}{l}{\textbf{\texttt{parallel}}} \\
\midrule
\texttt{sl\_1} & $L(N,P)=c_0+c_NN^{-\alpha}+c_PP^{-\beta}+c_{NP}N^{-\alpha}P^{-\beta}$ & 6 & 1.00 {\tiny $\pm$ 0.00} \\
\texttt{sl\_2} & $L(N,P)=c_0+c_NN^{-\alpha}+c_PP^{-\beta}$ & 5 & 1.00 {\tiny $\pm$ 0.00} \\
\texttt{sl\_3} & $L(N,P)=aN^b+\frac{c}{1+P}+d$ & 4 & 1.00 {\tiny $\pm$ 0.00} \\
\texttt{sl\_4} & $L(N,P)=aN^b+cP^{-1/2}+d$ & 4 & 1.00 {\tiny $\pm$ 0.00} \\
\texttt{sl\_5} & $L(N,P)=\left(\frac{A}{N(k\log P+1)}\right)^\alpha+E$ & 4 & 1.00 {\tiny $\pm$ 0.00} \\
\texttt{sl\_6} & $L(N,P)=c_0+c_1(N^{-\alpha}+P^{-\beta})$ & 4 & 1.00 {\tiny $\pm$ 0.00} \\
\texttt{sl\_7} & $L(N,P)=L_0+\frac{AN^{-\alpha}}{1+k\ln P}$ & 4 & 1.00 {\tiny $\pm$ 0.00} \\
\texttt{sl\_8} & $L(N,P)=\frac{aN^b+c}{1+d\log P}$ & 4 & 1.00 {\tiny $\pm$ 0.00} \\
\texttt{sl\_9} & $L(N,P)=AN^{-\alpha}P^{-\beta}$ & 3 & 1.00 {\tiny $\pm$ 0.00} \\
\texttt{sl\_10} & $L(N,P)=(AN^{-\alpha}+E)P^{-\beta}$ & 4 & 1.00 {\tiny $\pm$ 0.00} \\
\addlinespace
\midrule
\multicolumn{4}{l}{\textbf{\texttt{sparsity}}} \\
\midrule
\texttt{sl\_1} & $L(P,N_2)=e^{d_1}P^{-a}N_2^{-b}e^{c\log P\log N_2}+e^{d_3}$ & 5 & 0.28 {\tiny $\pm$ 0.00} \\
\texttt{sl\_2} & $L(P,N_2)=e^{d_1}P^{-a}N_2^{-b}+e^{d_3}$ & 4 & 0.52 {\tiny $\pm$ 0.00} \\
\texttt{sl\_3} & $L(P,N_2)=e^{d_1}P^{-a}+e^{d_2}N_2^{-b}e^{c\log P\log N_2}+e^{d_3}$ & 6 & 0.27 {\tiny $\pm$ 0.00} \\
\texttt{sl\_4} & $L(P,N_2)=e^{d_1}P^{-a}+e^{d_2}N_2^{-b}+e^{d_3}$ & 5 & 0.40 {\tiny $\pm$ 0.00} \\
\addlinespace
\midrule
\multicolumn{4}{l}{\textbf{\texttt{vocab}}} \\
\midrule
\texttt{sl\_1} & $L(N,V,D)=c_0+AV^bN^eD^g$ & 5 & 0.89 {\tiny $\pm$ 0.30} \\
\texttt{sl\_2} & $L(N,V,D)=L+A\,M_r(N^{-\alpha},D^{-\beta})\bigl(1+C(\log V-v_0)^2\bigr)$ & 7 & 0.99 {\tiny $\pm$ 0.00} \\
\texttt{sl\_3} & $L(N,V,D)=L_0+\left((aN^{-\alpha})^q+\bigl(b(DV^\phi)^{-\beta}\bigr)^q\right)^{1/q}$ & 7 & 0.99 {\tiny $\pm$ 0.00} \\
\texttt{sl\_4} & $L(N,V,D)=L_{\inf}+A\max(N^a,\lambda D^b)^{-d}V^{-g}$ & 7 & 0.83 {\tiny $\pm$ 0.19} \\
\texttt{sl\_5} & $L(N,V,D)=p_0N^{p_1}V^{p_2}D^{p_3}+p_4N^{p_5}+p_6$ & 7 & 0.99 {\tiny $\pm$ 0.00} \\
\texttt{sl\_6} & $L(N,V,D)=A(NV^{k_1})^{-\alpha}+B(DV^{k_2})^{-\beta}+c_0$ & 7 & 0.77 {\tiny $\pm$ 0.31} \\
\texttt{sl\_7} & $L(N,V,D)=AN^{-\alpha}D^{-\beta}+BV^\gamma D^{-\delta}+c_0$ & 7 & 0.98 {\tiny $\pm$ 0.02} \\
\texttt{sl\_8} & $L(N,V,D)=c_0+c_1\log V+V^\beta(c_2N^{-\alpha}+c_3D^{-\gamma})$ & 7 & 0.97 {\tiny $\pm$ 0.04} \\
\texttt{sl\_9} & $L(N,V,D)=AN^{-\alpha}D^{-\beta}(1+\gamma\log V)+\delta V^\epsilon+L_{\inf}$ & 7 & 0.98 {\tiny $\pm$ 0.00} \\
\texttt{sl\_10} & $L(N,V,D)=L_{\min}+\exp\!\bigl(a+b_P\log N+b_{V1}\log V+b_{V2}\log^2V+b_D\log D+b_{VD}\log V\log D\bigr)$ & 7 & 0.98 {\tiny $\pm$ 0.00} \\

\end{longtable}

\section{Basin Estimation and Posterior Approximation}
\label{app:basin_estimation}

We assume access to a set of locally optimal parameter vectors obtained by refitting the scaling law from multiple initializations on the current dataset $\mathcal D_t$. This appendix explains how these local solutions are converted into the basin approximation
\[
p(\theta\mid\mathcal D_t)\approx \sum_{k=1}^K w_k\,\mathcal N(\theta_k,\Sigma_k).
\]

\subsection{Local covariance approximation}

We assume access to a collection of locally optimal parameter vectors
\[
\{\tilde\theta_m\}_{m=1}^M
\]
obtained by refitting the scaling law from multiple initializations on the current dataset $\mathcal D_t$. This subsection explains how each local solution is converted into a local Gaussian approximation, which will later be consolidated into the basin mixture
\[
p(\theta\mid\mathcal D_t)
\approx
\sum_{k=1}^K w_k\,\mathcal N(\theta_k,\Sigma_k).
\]

For a local optimum $\tilde\theta_m$, let
\[
\mathcal L_t(\theta)
=
\frac{1}{|\mathcal D_t|}
\sum_{(x_i,y_i)\in\mathcal D_t}
\|f(x_i;\theta)-y_i\|_2^2
\]
denote the empirical mean-squared error on the currently observed data. Around $\tilde\theta_m$, we approximate the local curvature of this objective using a Gauss--Newton/Fisher-style matrix. Concretely, let
\[
J_m
=
J(\tilde\theta_m;\mathcal D_t)
\]
be the Jacobian of model predictions with respect to the parameters, evaluated on all observations in $\mathcal D_t$ and flattened across data points (and output dimensions, when applicable). Given the current noise variance estimate $\sigma^2$, we define
\[
H_m
=
\frac{1}{\sigma^2}J_m^\top J_m + \Lambda_{\mathrm{prior}},
\]
where $\Lambda_{\mathrm{prior}}$ is a diagonal prior-precision matrix used to stabilize weakly identified directions. We then approximate the local parameter uncertainty around $\tilde\theta_m$ by
\[
\tilde\Sigma_m
=
H_m^{-1}.
\]

This yields the local Gaussian approximation
\[
q_m(\theta)
=
\mathcal N(\tilde\theta_m,\tilde\Sigma_m)
\]
for each candidate local optimum.

For parameters constrained to be positive, we additionally rescale the corresponding Jacobian columns by the current parameter values before forming $H_m$. Equivalently, this amounts to measuring local sensitivity in a log-parameterization for those coordinates, which improves numerical stability when different parameters operate on very different scales. In implementation, the inversion of $H_m$ is further stabilized by standard spectral regularization, including eigenvalue flooring and small diagonal correction when necessary.

At this stage, each local refit $\tilde\theta_m$ is associated with a Gaussian approximation $q_m(\theta)$. These candidate components are not yet the final basins in the mixture posterior, because several local optima may induce nearly identical predictive behavior. The next subsection therefore clusters these local solutions in prediction space and consolidates them into a smaller set of representative basins.

\subsection{Prediction-space clustering of local optima}

The local Gaussian approximations constructed above are still over-complete: different local optima may correspond to essentially the same extrapolative behavior, even when their parameter values differ substantially. This is particularly common for nonlinear scaling laws with weakly identifiable directions. Since our downstream objective is prediction accuracy on a target region rather than parameter recovery itself, we consolidate local optima in prediction space rather than in parameter space.

For each local solution $\tilde\theta_m$ with covariance $\tilde\Sigma_m$, we evaluate its induced predictive distribution on an evaluation set $\mathcal X_{\mathrm{eval}}$. In our implementation, $\mathcal X_{\mathrm{eval}}$ is chosen to be the target region used by the acquisition function. Under the local linear approximation, the predictive distribution at a point $x \in \mathcal X_{\mathrm{eval}}$ is
\[
f(x;\theta)\mid \theta \sim q_m
\;\approx\;
\mathcal N\!\bigl(\mu_m(x),\,v_m(x)\bigr),
\]
where
\[
\mu_m(x)=f(x;\tilde\theta_m),
\qquad
v_m(x)=J_x(\tilde\theta_m)\,\tilde\Sigma_m\,J_x(\tilde\theta_m)^\top+\sigma^2,
\]
and $J_x(\tilde\theta_m)=\left.\frac{\partial f(x;\theta)}{\partial\theta}\right|_{\theta=\tilde\theta_m}$ is the parameter Jacobian at $x$.

We then measure the discrepancy between two local optima $\tilde\theta_m$ and $\tilde\theta_n$ by comparing their predictive Gaussians across $\mathcal X_{\mathrm{eval}}$. For scalar outputs, the pointwise symmetric KL divergence between
\[
\mathcal N(\mu_m(x),v_m(x))
\quad\text{and}\quad
\mathcal N(\mu_n(x),v_n(x))
\]
is
\[
\mathrm{SKL}_{m,n}(x)
=
\frac14\left(
\frac{v_m(x)}{v_n(x)}
+
\frac{v_n(x)}{v_m(x)}
-2
+
(\mu_m(x)-\mu_n(x))^2
\left(
\frac{1}{v_m(x)}+\frac{1}{v_n(x)}
\right)
\right).
\]
Averaging over the evaluation region gives the prediction-space dissimilarity
\[
d_{mn}
=
\frac{1}{|\mathcal X_{\mathrm{eval}}|}
\sum_{x\in\mathcal X_{\mathrm{eval}}}
\mathrm{SKL}_{m,n}(x).
\]

Using the pairwise dissimilarity matrix $\{d_{mn}\}$, we perform agglomerative hierarchical clustering to group local optima with similar predictive behavior. The clustering threshold is selected data-adaptively by maximizing the silhouette score over candidate cuts of the hierarchy. This procedure yields a partition of the local optima into clusters
\[
\mathcal C_1,\dots,\mathcal C_K,
\]
which we interpret as the final basins.

For each cluster $\mathcal C_k$, we choose a single representative local optimum as the basin center. Specifically, we select the member with the smallest empirical fitting error on the current dataset:
\[
m_k
=
\arg\min_{m\in\mathcal C_k}
\mathcal L_t(\tilde\theta_m),
\qquad
\theta_k
=
\tilde\theta_{m_k}.
\]
Its associated covariance approximation is inherited from the same representative:
\[
\Sigma_k
=
\tilde\Sigma_{m_k}.
\]

The output of this step is therefore a reduced collection of representative basins
\[
\{(\theta_k,\Sigma_k)\}_{k=1}^K,
\]
which preserves distinct extrapolative behaviors while removing redundant local optima. The remaining ingredient is to assign mixture weights to these representative basins; this is described in the next subsection.

\subsection{Representative basins and mixture weights}

After clustering the local optima in prediction space, we obtain a reduced set of representative basins
\[
\{(\theta_k,\Sigma_k)\}_{k=1}^K,
\]
where each representative parameter $\theta_k$ is selected from one prediction-space cluster, and $\Sigma_k$ is the corresponding local covariance approximation inherited from that representative.

To complete the mixture approximation, we associate each basin with a weight
\[
w_k \approx p(B=k \mid \mathcal D_t),
\]
where $B\in\{1,\dots,K\}$ is a latent basin indicator specifying which basin generated the current local posterior approximation. Thus, $w_k$ represents the posterior probability that basin $k$ is the relevant mode given the observations collected so far. In practice, exact computation of $p(B=k\mid\mathcal D_t)$ is generally intractable, so we approximate it using a basin-level evidence score. We consider two natural choices. 

\paragraph{Option 1: BIC-style approximation.}
A simple approximation is to rank basins by an information criterion derived from their empirical fit on the current dataset. Let
\[
\mathrm{MSE}_k
=
\mathcal L_t(\theta_k)
=
\frac{1}{|\mathcal D_t|}
\sum_{(x_i,y_i)\in\mathcal D_t}
\|f(x_i;\theta_k)-y_i\|_2^2
\]
and let $n_{\mathrm{obs}}=|\mathcal D_t|$. We define
\[
\mathrm{BIC}_k
=
n_{\mathrm{obs}}\log(\mathrm{MSE}_k)+p\log n_{\mathrm{obs}},
\]
where $p$ is the number of free parameters. The basin probabilities are then approximated by
\[
w_k
=
\frac{
\exp\!\left(-\frac{\mathrm{BIC}_k}{2T}\right)
}{
\sum_{\ell=1}^K
\exp\!\left(-\frac{\mathrm{BIC}_\ell}{2T}\right)
},
\]
where $T>0$ is a temperature parameter.

\paragraph{Option 2: Laplace-approximate basin posterior.}
A more direct approximation is obtained by locally approximating the contribution of each basin to the posterior normalizing constant. Let $H_k$ denote the local curvature matrix at $\theta_k$, with $\Sigma_k \approx H_k^{-1}$. Then, up to a common normalization constant,
\[
p(B=k\mid\mathcal D_t)
\;\propto\;
\pi(\theta_k)\,
\exp\!\left(
-\frac{|\mathcal D_t|}{2\sigma^2}\mathcal L_t(\theta_k)
\right)
|H_k|^{-1/2},
\]
which yields the normalized weight
\[
w_k
=
\frac{
\pi(\theta_k)\,
\exp\!\left(
-\frac{|\mathcal D_t|}{2\sigma^2}\mathcal L_t(\theta_k)
\right)
|H_k|^{-1/2}
}{
\sum_{\ell=1}^K
\pi(\theta_\ell)\,
\exp\!\left(
-\frac{|\mathcal D_t|}{2\sigma^2}\mathcal L_t(\theta_\ell)
\right)
|H_\ell|^{-1/2}
}.
\]

In our experiments, we use the BIC-style approximation for robustness, while the Laplace form provides a more principled local-evidence interpretation of the same quantity.

\section{Derivation of the Acquisition Function}
\label{app:utility_derivation}

This appendix derives the target-aware acquisition function used in the main text. Starting from the basin-mixture approximation
\[
p(\theta\mid\mathcal D_t)
\approx
\sum_{k=1}^K w_k\,\mathcal N(\theta_k,\Sigma_k),
\]
we show how the target-region uncertainty measure
\[
\mathrm{MSPE}_{\mathrm{tar}}
=
\frac{1}{|\mathcal X_{\mathrm{tar}}|}
\mathbb E_{\theta\sim p(\theta\mid\mathcal D_t)}
\bigl[
\|F(\theta)-\bar f\|_2^2
\bigr]
\]
decomposes into intra-basin and inter-basin terms, and how this decomposition leads to the candidate utility
\[
\Delta \mathrm{MSPE}_{\mathrm{tar}}(x)
=
\Delta V_{\mathrm{intra}}(x)
+
\Delta V_{\mathrm{inter}}(x).
\]
Throughout, the key approximation is local: within each basin, we linearize the predictor around its representative parameter $\theta_k$, while retaining the multimodal mixture structure across basins.

\subsection{Local linearization within each basin}

Fix a basin $k$ with local posterior approximation
\[
\theta \mid (B=k,\mathcal D_t)
\approx
\mathcal N(\theta_k,\Sigma_k).
\]
To obtain tractable expressions for predictive uncertainty and posterior updates after adding a new observation, we linearize the scaling law around the basin representative $\theta_k$.

For the target region $\mathcal X_{\mathrm{tar}}$, recall the prediction map
\[
F(\theta)
=
\bigl(f(x;\theta)\bigr)_{x\in\mathcal X_{\mathrm{tar}}}
\in \mathbb R^{|\mathcal X_{\mathrm{tar}}|}.
\]
A first-order Taylor expansion around $\theta_k$ gives
\[
F(\theta)
\approx
F(\theta_k)
+
J_k(\theta-\theta_k),
\]
where
\[
J_k
=
\left.
\frac{\partial F(\theta)}{\partial \theta}
\right|_{\theta=\theta_k}
\in
\mathbb R^{|\mathcal X_{\mathrm{tar}}|\times p}.
\]
Writing
\[
\hat f_k = F(\theta_k),
\]
we obtain the Gaussian approximation
\[
F(\theta)\mid (B=k,\mathcal D_t)
\approx
\mathcal N\!\bigl(\hat f_k,\; J_k\Sigma_kJ_k^\top\bigr).
\]

Similarly, for a candidate experiment $x \in \mathcal X_{\mathrm{cand}}$, define the scalar predictive mean
\[
m_k(x)=f(x;\theta_k)
\]
and the parameter Jacobian
\[
j_k(x)
=
\left.
\frac{\partial f(x;\theta)}{\partial\theta}
\right|_{\theta=\theta_k}
\in \mathbb R^{p}.
\]
Then the same first-order expansion yields
\[
f(x;\theta)
\approx
m_k(x)+j_k(x)^\top(\theta-\theta_k),
\]
so under the local Gaussian approximation,
\[
y\mid (x,B=k,\mathcal D_t)
\approx
\mathcal N\!\bigl(m_k(x),\, s_k^2(x)\bigr),
\qquad
s_k^2(x)
=
\sigma^2+j_k(x)^\top\Sigma_k j_k(x).
\]

The corresponding posterior update within basin $k$ remains Gaussian. After observing $(x,y)$, the updated covariance is given by the standard rank-one linear-Gaussian update
\[
\Sigma_k^{+}(x)
=
\Sigma_k
-
\frac{
\Sigma_k j_k(x)j_k(x)^\top \Sigma_k
}{
s_k^2(x)
},
\]
and the posterior mean of the target-region prediction vector updates as
\[
\hat f_k^{+}(x,y)
=
\hat f_k
+
g_k(x)\bigl(y-m_k(x)\bigr),
\qquad
g_k(x)
=
\frac{J_k\Sigma_k j_k(x)}{s_k^2(x)}
\in
\mathbb R^{|\mathcal X_{\mathrm{tar}}|}.
\]

These local update formulas are the basic ingredients for the derivations below: $\Sigma_k^{+}(x)$ determines the reduction in within-basin predictive variance, while $\hat f_k^{+}(x,y)$ determines how a new observation changes the disagreement between basins.

\subsection{Decomposition of target-region MSPE}

We now derive the decomposition of the target-region uncertainty objective used in the main text. Recall that
\[
\mathrm{MSPE}_{\mathrm{tar}}
=
\frac{1}{|\mathcal X_{\mathrm{tar}}|}
\mathbb E_{\theta\sim p(\theta\mid\mathcal D_t)}
\bigl[
\|F(\theta)-\bar f\|_2^2
\bigr],
\qquad
\bar f
=
\mathbb E_{\theta\sim p(\theta\mid\mathcal D_t)}[F(\theta)].
\]
Introducing the latent basin indicator $B\in\{1,\dots,K\}$, the mixture approximation can be written as
\[
p(\theta\mid\mathcal D_t)
\approx
\sum_{k=1}^K
p(B=k\mid\mathcal D_t)\,
p(\theta\mid B=k,\mathcal D_t),
\]
with
\[
p(B=k\mid\mathcal D_t)\approx w_k,
\qquad
p(\theta\mid B=k,\mathcal D_t)\approx \mathcal N(\theta_k,\Sigma_k).
\]
Under the local linearization from the previous subsection,
\[
F(\theta)\mid (B=k,\mathcal D_t)
\approx
\mathcal N\!\bigl(\hat f_k,\;J_k\Sigma_kJ_k^\top\bigr),
\qquad
\hat f_k = F(\theta_k).
\]

The target-region MSPE is simply the total variance of the random vector $F(\theta)$, normalized by $|\mathcal X_{\mathrm{tar}}|$. Applying the law of total variance with respect to the basin indicator $B$ gives
\[
\mathrm{MSPE}_{\mathrm{tar}}
=
\frac{1}{|\mathcal X_{\mathrm{tar}}|}
\left(
\mathbb E_B\!\left[\operatorname{tr}\!\bigl(\operatorname{Cov}(F(\theta)\mid B,\mathcal D_t)\bigr)\right]
+
\operatorname{tr}\!\bigl(\operatorname{Cov}(\mathbb E[F(\theta)\mid B,\mathcal D_t])\bigr)
\right).
\]
We evaluate the two terms separately.

For the first term, conditioning on basin $k$ yields
\[
\operatorname{Cov}(F(\theta)\mid B=k,\mathcal D_t)
\approx
J_k\Sigma_kJ_k^\top,
\]
so
\[
\mathbb E_B\!\left[\operatorname{tr}\!\bigl(\operatorname{Cov}(F(\theta)\mid B,\mathcal D_t)\bigr)\right]
\approx
\sum_{k=1}^K
w_k\,\operatorname{tr}(J_k\Sigma_kJ_k^\top).
\]

For the second term, the conditional mean is
\[
\mathbb E[F(\theta)\mid B=k,\mathcal D_t]
\approx
\hat f_k,
\]
and the global posterior mean is therefore
\[
\bar f
=
\mathbb E[F(\theta)\mid\mathcal D_t]
\approx
\sum_{k=1}^K w_k \hat f_k.
\]
Hence,
\[
\operatorname{tr}\!\bigl(\operatorname{Cov}(\mathbb E[F(\theta)\mid B,\mathcal D_t])\bigr)
=
\sum_{k=1}^K
w_k\,\|\hat f_k-\bar f\|_2^2.
\]

Combining the two terms yields
\[
\mathrm{MSPE}_{\mathrm{tar}}
=
\frac{1}{|\mathcal X_{\mathrm{tar}}|}
\sum_{k=1}^K
w_k\,\operatorname{tr}(J_k\Sigma_kJ_k^\top)
+
\frac{1}{|\mathcal X_{\mathrm{tar}}|}
\sum_{k=1}^K
w_k\,\|\hat f_k-\bar f\|_2^2.
\]
We therefore define
\[
V_{\mathrm{intra}}
=
\frac{1}{|\mathcal X_{\mathrm{tar}}|}
\sum_{k=1}^K
w_k\,\operatorname{tr}(J_k\Sigma_kJ_k^\top),
\qquad
V_{\mathrm{inter}}
=
\frac{1}{|\mathcal X_{\mathrm{tar}}|}
\sum_{k=1}^K
w_k\,\|\hat f_k-\bar f\|_2^2,
\]
so that
\[
\mathrm{MSPE}_{\mathrm{tar}}=V_{\mathrm{intra}}+V_{\mathrm{inter}}.
\]

The term $V_{\mathrm{intra}}$ measures the average predictive variance that remains within each basin after conditioning on which local mode is correct, while $V_{\mathrm{inter}}$ measures the residual disagreement between basin-level extrapolations. This decomposition is the basis for our acquisition function: a useful new experiment should either reduce uncertainty within a plausible basin or help distinguish between basins that extrapolate differently.

\subsection{Derivation of the intra-basin utility}

We now derive the reduction in the within-basin term
\[
V_{\mathrm{intra}}
=
\frac{1}{|\mathcal X_{\mathrm{tar}}|}
\sum_{k=1}^K
w_k\,\operatorname{tr}(J_k\Sigma_kJ_k^\top)
\]
after querying a candidate experiment $x$.

After observing an outcome $y$ at $x$, both the basin posterior probabilities and the within-basin covariances are updated. The exact updated intra-basin term is therefore
\[
V_{\mathrm{intra}}^{+}(x,y)
=
\frac{1}{|\mathcal X_{\mathrm{tar}}|}
\sum_{k=1}^K
w_k^{+}(x,y)\,
\operatorname{tr}\!\bigl(J_k\Sigma_k^{+}(x)J_k^\top\bigr),
\]
where
\[
w_k^{+}(x,y)
=
p(B=k\mid x,y,\mathcal D_t)
\]
and
\[
\Sigma_k^{+}(x)
=
\Sigma_k
-
\frac{
\Sigma_k j_k(x)j_k(x)^\top \Sigma_k
}{
s_k^2(x)
},
\qquad
s_k^2(x)=\sigma^2+j_k(x)^\top\Sigma_k j_k(x).
\]
Note that $\Sigma_k^{+}(x)$ depends on the queried location $x$ but not on the realized value $y$.

The intra-basin utility is defined by
\[
\Delta V_{\mathrm{intra}}(x)
=
V_{\mathrm{intra}}
-
\mathbb E_{y\mid x,\mathcal D_t}
\bigl[
V_{\mathrm{intra}}^{+}(x,y)
\bigr].
\]
Substituting the expression above gives
\[
\mathbb E_{y\mid x,\mathcal D_t}
\bigl[
V_{\mathrm{intra}}^{+}(x,y)
\bigr]
=
\frac{1}{|\mathcal X_{\mathrm{tar}}|}
\sum_{k=1}^K
\mathbb E_{y\mid x,\mathcal D_t}
\bigl[
w_k^{+}(x,y)
\bigr]
\operatorname{tr}\!\bigl(J_k\Sigma_k^{+}(x)J_k^\top\bigr).
\]
It remains to evaluate the expectation of the updated basin weights.

By Bayes' rule,
\[
w_k^{+}(x,y)
=
\frac{
w_k\,p(y\mid x,B=k,\mathcal D_t)
}{
p(y\mid x,\mathcal D_t)
}.
\]
Therefore,
\[
\begin{aligned}
\mathbb E_{y\mid x,\mathcal D_t}[w_k^{+}(x,y)]
&=
\int
\frac{
w_k\,p(y\mid x,B=k,\mathcal D_t)
}{
p(y\mid x,\mathcal D_t)
}
\,p(y\mid x,\mathcal D_t)\,dy \\
&=
w_k\int p(y\mid x,B=k,\mathcal D_t)\,dy
=
w_k.
\end{aligned}
\]
Hence,
\[
\mathbb E_{y\mid x,\mathcal D_t}
\bigl[
V_{\mathrm{intra}}^{+}(x,y)
\bigr]
=
\frac{1}{|\mathcal X_{\mathrm{tar}}|}
\sum_{k=1}^K
w_k\,\operatorname{tr}\!\bigl(J_k\Sigma_k^{+}(x)J_k^\top\bigr),
\]
and thus
\[
\Delta V_{\mathrm{intra}}(x)
=
\frac{1}{|\mathcal X_{\mathrm{tar}}|}
\sum_{k=1}^K
w_k
\left[
\operatorname{tr}(J_k\Sigma_kJ_k^\top)
-
\operatorname{tr}(J_k\Sigma_k^{+}(x)J_k^\top)
\right].
\]

Substituting the rank-one update for $\Sigma_k^{+}(x)$ yields
\[
\Delta V_{\mathrm{intra}}(x)
=
\frac{1}{|\mathcal X_{\mathrm{tar}}|}
\sum_{k=1}^K
w_k\,
\operatorname{tr}\!\left(
J_k
\frac{
\Sigma_k j_k(x)j_k(x)^\top \Sigma_k
}{
s_k^2(x)
}
J_k^\top
\right).
\]
Using cyclic invariance of the trace,
\[
\operatorname{tr}\!\left(
J_k\Sigma_k j_k(x)j_k(x)^\top \Sigma_k J_k^\top
\right)
=
j_k(x)^\top \Sigma_k J_k^\top J_k \Sigma_k j_k(x),
\]
so
\[
\Delta V_{\mathrm{intra}}(x)
=
\frac{1}{|\mathcal X_{\mathrm{tar}}|}
\sum_{k=1}^K
w_k\,
\frac{
j_k(x)^\top \Sigma_k J_k^\top J_k \Sigma_k j_k(x)
}{
s_k^2(x)
}.
\]
Equivalently,
\[
\Delta V_{\mathrm{intra}}(x)
=
\frac{1}{|\mathcal X_{\mathrm{tar}}|}
\sum_{k=1}^K
w_k\,
\frac{
\|J_k\Sigma_k j_k(x)\|_2^2
}{
\sigma^2 + j_k(x)^\top\Sigma_k j_k(x)
}.
\]

\subsection{Derivation of the inter-basin utility}

We now derive the reduction in the between-basin term
\[
V_{\mathrm{inter}}
=
\frac{1}{|\mathcal X_{\mathrm{tar}}|}
\sum_{k=1}^K
w_k\,\|\hat f_k-\bar f\|_2^2,
\qquad
\bar f = \sum_{k=1}^K w_k \hat f_k,
\]
after querying a candidate experiment $x$.

Unlike the intra-basin term, the inter-basin term depends on the relative positions and weights of the basin-level predictions. After observing an outcome $y$ at $x$, both quantities change: the posterior probability of each basin is updated, and within each basin the target-region prediction mean is shifted by the new observation.

\paragraph{Updated basin weights.}
By Bayes' rule, the posterior probability of basin $k$ after observing $(x,y)$ is
\[
w_k^{+}(x,y)
=
p(B=k\mid x,y,\mathcal D_t)
=
\frac{
w_k\,p(y\mid x,B=k,\mathcal D_t)
}{
p(y\mid x,\mathcal D_t)
}.
\]
Under the local linear-Gaussian approximation,
\[
p(y\mid x,B=k,\mathcal D_t)
\approx
\phi\!\left(y;\,m_k(x),\,s_k^2(x)\right),
\]
where
\[
m_k(x)=f(x;\theta_k),
\qquad
s_k^2(x)=\sigma^2+j_k(x)^\top\Sigma_k j_k(x),
\]
and $\phi(\,\cdot\,;\mu,\nu)$ denotes the Gaussian density with mean $\mu$ and variance $\nu$. Therefore,
\[
w_k^{+}(x,y)
=
\frac{
w_k\,\phi\!\left(y;\,m_k(x),\,s_k^2(x)\right)
}{
\sum_{\ell=1}^K
w_\ell\,\phi\!\left(y;\,m_\ell(x),\,s_\ell^2(x)\right)
}.
\]

\paragraph{Updated basin-level target predictions.}
Within basin $k$, the posterior mean of the target-region prediction vector updates according to
\[
\hat f_k^{+}(x,y)
=
\hat f_k
+
g_k(x)\bigl(y-m_k(x)\bigr),
\qquad
g_k(x)
=
\frac{J_k\Sigma_k j_k(x)}{s_k^2(x)}
\in
\mathbb R^{|\mathcal X_{\mathrm{tar}}|}.
\]
Thus the updated global posterior mean on the target region is
\[
\bar f^{+}(x,y)
=
\sum_{k=1}^K
w_k^{+}(x,y)\,\hat f_k^{+}(x,y).
\]

\paragraph{Updated inter-basin uncertainty.}
Conditioned on the new observation $(x,y)$, the between-basin term becomes
\[
V_{\mathrm{inter}}^{+}(x,y)
=
\frac{1}{|\mathcal X_{\mathrm{tar}}|}
\sum_{k=1}^K
w_k^{+}(x,y)\,
\left\|
\hat f_k^{+}(x,y)-\bar f^{+}(x,y)
\right\|_2^2.
\]
Accordingly, the inter-basin utility of candidate $x$ is
\[
\Delta V_{\mathrm{inter}}(x)
=
V_{\mathrm{inter}}
-
\mathbb E_{y\mid x,\mathcal D_t}
\bigl[
V_{\mathrm{inter}}^{+}(x,y)
\bigr].
\]

The predictive distribution of $y$ under the current mixture is
\[
p(y\mid x,\mathcal D_t)
=
\sum_{k=1}^K
w_k\,\phi\!\left(y;\,m_k(x),\,s_k^2(x)\right),
\]
so the expectation above can be written explicitly as the one-dimensional integral
\[
\mathbb E_{y\mid x,\mathcal D_t}
\bigl[
V_{\mathrm{inter}}^{+}(x,y)
\bigr]
=
\int
V_{\mathrm{inter}}^{+}(x,y)\,
p(y\mid x,\mathcal D_t)\,dy.
\]

In contrast to the intra-basin case, this expectation does not collapse to a simpler weighted average, because both the updated responsibilities $w_k^{+}(x,y)$ and the updated basin predictions $\hat f_k^{+}(x,y)$ depend nonlinearly on the realized outcome $y$. Intuitively, a candidate receives high inter-basin utility when different basins predict substantially different outcomes at $x$, so that observing $y$ is likely to either reweight the basin probabilities or pull their target-region predictions closer together. In the next subsection, we show that this integral admits an efficient pairwise form that can be evaluated by one-dimensional numerical quadrature.

\subsection{Pairwise form and one-dimensional quadrature}

A convenient identity for the between-basin variance is
\[
\sum_{k=1}^K w_k \,\|\hat f_k-\bar f\|_2^2
=
\sum_{1\le k<\ell\le K}
w_k w_\ell \,\|\hat f_k-\hat f_\ell\|_2^2.
\]
Applying this identity to the updated basin mixture yields
\[
V_{\mathrm{inter}}^{+}(x,y)
=
\frac{1}{|\mathcal X_{\mathrm{tar}}|}
\sum_{1\le k<\ell\le K}
w_k^{+}(x,y)\,w_\ell^{+}(x,y)\,
\left\|
\hat f_k^{+}(x,y)-\hat f_\ell^{+}(x,y)
\right\|_2^2.
\]
This pairwise form is more convenient than the centered form because it separates the contribution of each basin pair and avoids recomputing the global mean explicitly.

Using the linear update
\[
\hat f_k^{+}(x,y)
=
\hat f_k+g_k(x)\bigl(y-m_k(x)\bigr),
\]
the difference between two updated basin means is
\[
\hat f_k^{+}(x,y)-\hat f_\ell^{+}(x,y)
=
\underbrace{
\hat f_k-\hat f_\ell-g_k(x)m_k(x)+g_\ell(x)m_\ell(x)
}_{a_{k\ell}(x)}
+
\underbrace{
\bigl(g_k(x)-g_\ell(x)\bigr)
}_{b_{k\ell}(x)}
\,y.
\]
Hence, for each pair $(k,\ell)$,
\[
\left\|
\hat f_k^{+}(x,y)-\hat f_\ell^{+}(x,y)
\right\|_2^2
=
\|a_{k\ell}(x)+b_{k\ell}(x)y\|_2^2,
\]
which is a quadratic polynomial in $y$:
\[
\left\|
\hat f_k^{+}(x,y)-\hat f_\ell^{+}(x,y)
\right\|_2^2
=
A_{k\ell}(x)+B_{k\ell}(x)\,y+C_{k\ell}(x)\,y^2,
\]
with coefficients
\[
A_{k\ell}(x)=\|a_{k\ell}(x)\|_2^2,
\qquad
B_{k\ell}(x)=2\,a_{k\ell}(x)^\top b_{k\ell}(x),
\qquad
C_{k\ell}(x)=\|b_{k\ell}(x)\|_2^2.
\]

Next, recall that the updated basin weights satisfy
\[
w_k^{+}(x,y)
=
\frac{
w_k\,\phi_k(y;x)
}{
\sum_{r=1}^K w_r\,\phi_r(y;x)
},
\qquad
\phi_k(y;x)
=
\phi\!\left(y;\,m_k(x),\,s_k^2(x)\right).
\]
Therefore, for any pair $(k,\ell)$,
\[
w_k^{+}(x,y)\,w_\ell^{+}(x,y)\,p(y\mid x,\mathcal D_t)
=
\frac{
w_k w_\ell\,\phi_k(y;x)\phi_\ell(y;x)
}{
\sum_{r=1}^K w_r\,\phi_r(y;x)
}.
\]
Substituting this into the expectation of the updated inter-basin term gives
\begin{align*}
    &\mathbb E_{y\mid x,\mathcal D_t}
\bigl[
V_{\mathrm{inter}}^{+}(x,y)
\bigr]\\
=&
\frac{1}{|\mathcal X_{\mathrm{tar}}|}
\sum_{1\le k<\ell\le K}
w_k w_\ell
\int
\frac{
\phi_k(y;x)\phi_\ell(y;x)
}{
\sum_{r=1}^K w_r\,\phi_r(y;x)
}
\left(
A_{k\ell}(x)+B_{k\ell}(x)\,y+C_{k\ell}(x)\,y^2
\right)
\mathrm{d}y.
\end{align*}

This is the form used in our implementation. For a fixed candidate $x$, the expectation reduces to a one-dimensional integral over the scalar observation $y$, and the only dependence on the target region enters through the precomputed coefficient vectors $a_{k\ell}(x)$ and $b_{k\ell}(x)$. Consequently, the inter-basin utility can be evaluated efficiently by numerical quadrature even when the target region contains many points.

Finally, combining this expression with
\[
V_{\mathrm{inter}}
=
\frac{1}{|\mathcal X_{\mathrm{tar}}|}
\sum_{1\le k<\ell\le K}
w_k w_\ell \,\|\hat f_k-\hat f_\ell\|_2^2
\]
gives
\[
\Delta V_{\mathrm{inter}}(x)
=
V_{\mathrm{inter}}
-
\mathbb E_{y\mid x,\mathcal D_t}
\bigl[
V_{\mathrm{inter}}^{+}(x,y)
\bigr].
\]
In practice, we evaluate the one-dimensional integral numerically on a finite grid covering the predictive support of the current basin mixture at the candidate point.

\subsection{Final cost-aware acquisition score}

Combining the two components derived above, the total expected reduction in target-region uncertainty from querying candidate $x$ is
\[
\Delta \mathrm{MSPE}_{\mathrm{tar}}(x)
=
\Delta V_{\mathrm{intra}}(x)
+
\Delta V_{\mathrm{inter}}(x).
\]
The intra-basin term captures how much the new observation is expected to reduce local predictive variance within each plausible basin, while the inter-basin term captures how much it is expected to reduce disagreement across basins.

To account for heterogeneous experiment costs, we rank candidates using the cost-aware score
\[
S(x)
=
\frac{
\Delta \mathrm{MSPE}_{\mathrm{tar}}(x)
}{
c(x)^\alpha
}
=
\frac{
\Delta V_{\mathrm{intra}}(x)+\Delta V_{\mathrm{inter}}(x)
}{
c(x)^\alpha
},
\]
where $c(x)$ is the cost of running experiment $x$, and $\alpha\ge 0$ controls the strength of cost penalization.

This form has a natural interpretation. When basin ambiguity is large, the inter-basin term tends to dominate, so the design prefers experiments that distinguish between qualitatively different extrapolations. Once the correct basin is largely identified, the acquisition increasingly behaves like a target-aware local optimal-design criterion, favoring experiments that most reduce predictive variance on $\mathcal X_{\mathrm{tar}}$ per unit cost. This is exactly the behavior desired in budget-constrained scaling-law fitting: early experiments should resolve global ambiguity, while later experiments should refine the locally relevant scaling trend.
\section{Evaluation Details and Robustness Analyses}
\label{app:additional_experiments}

This appendix collects additional robustness and implementation details for the \method{} sequential design procedure, including sensitivity to the cost exponent, robustness to the warm-start size, the cost of the warm-start phase, and acquisition-scoring complexity.

\subsection{Sensitivity to the Cost Exponent}
\label{app:alpha_sensitivity}

The cost exponent $\alpha$ controls the strength of cost penalization in the acquisition score. We use $\alpha=0.4$ as a single global value for all tasks and scaling-law instances, rather than tuning it per task. Table~\ref{tab:alpha_sensitivity} evaluates $\alpha \in \{0,0.25,0.4,0.5,0.75,1.0\}$ and shows that the method is stable across a broad middle range. The extreme settings behave as expected: $\alpha=0$ can spend budget too aggressively on expensive points in the low-budget regime, while $\alpha=1$ can over-prioritize cheap points and underperform at larger budgets.

\begin{table}[h]
\centering
\small
\setlength{\tabcolsep}{4.5pt}
\renewcommand{\arraystretch}{1.10}
\resizebox{\linewidth}{!}{%
\begin{tabular}{lcccccc}
\toprule
\textbf{Budget} & $\boldsymbol{\alpha=0}$ & $\boldsymbol{\alpha=0.25}$ & $\boldsymbol{\alpha=0.4}$ & $\boldsymbol{\alpha=0.5}$ & $\boldsymbol{\alpha=0.75}$ & $\boldsymbol{\alpha=1}$ \\
\midrule
$1\%$  & $-0.350 \pm 0.389$ & $\mathbf{0.288 \pm 0.317}$ & $0.271 \pm 0.439$ & $0.212 \pm 0.346$ & $0.234 \pm 0.352$ & $0.117 \pm 0.351$ \\
$5\%$  & $0.357 \pm 0.166$ & $\mathbf{0.694 \pm 0.090}$ & $0.683 \pm 0.091$ & $0.645 \pm 0.116$ & $0.614 \pm 0.175$ & $0.553 \pm 0.249$ \\
$10\%$ & $0.756 \pm 0.057$ & $\mathbf{0.796 \pm 0.055}$ & $0.786 \pm 0.065$ & $0.762 \pm 0.086$ & $0.718 \pm 0.135$ & $0.682 \pm 0.204$ \\
\bottomrule
\end{tabular}
}
\caption{\textbf{Cost-exponent sensitivity.} Sensitivity of \method{} to the cost-penalty exponent $\alpha$, aggregated over benchmark tasks and scaling-law instances. Higher target-region $R^2$ is better.}
\label{tab:alpha_sensitivity}
\end{table}

\subsection{Warm-Start Cost and Size}
\label{app:warm_start_analysis}

The design-based methods require an initial nonlinear fit before local Jacobian and covariance information can be used. We therefore initialize \textbf{D-opt}, \textbf{V-opt}, and \method{} with the same $2.5p$ lowest-cost observations, where $p$ is the number of law parameters, and count this initialization toward the total budget. Table~\ref{tab:warm_start_cost_fraction} reports the fraction of budget consumed by this warm start. The fraction is usually tiny relative to the full pool, though it is larger for tasks with more uniform costs and small candidate pools.

\begin{table}[h]
\centering
\small
\setlength{\tabcolsep}{3.0pt}
\renewcommand{\arraystretch}{1.10}
\resizebox{\linewidth}{!}{%
\begin{tabular}{lcccccccc}
\toprule
\textbf{Metric} & \textbf{lr\&bsz} & \textbf{domain} & \textbf{vocab} & \textbf{parallel} & \textbf{moe} & \textbf{data\_con} & \textbf{sparsity} & \textbf{farseer} \\
\midrule
Warm-start cost / full pool & $0.023\%$ & $5.000\%$ & $0.001\%$ & $1.615\%$ & $0.020\%$ & $0.007\%$ & $1.816\%$ & $0.028\%$ \\
Warm-start cost / final checkpoint & $0.23\%$ & $10.00\%$ & $0.01\%$ & $16.15\%$ & $0.20\%$ & $0.07\%$ & $3.63\%$ & $0.28\%$ \\
\bottomrule
\end{tabular}
}
\caption{\textbf{Warm-start cost fraction.} Cost of the shared $2.5p$ warm-start phase. The final checkpoint is $10\%$ of the full pool for most tasks and $50\%$ for \texttt{domain} and \texttt{sparsity}.}
\label{tab:warm_start_cost_fraction}
\end{table}

Table~\ref{tab:warm_start_size_ablation} varies the number of warm-start observations. The results are similar for $p$, $2.5p$, and $5p$, suggesting that performance is driven mainly by the acquisition strategy rather than by the precise initialization size.

\begin{table}[h]
\centering
\small
\setlength{\tabcolsep}{5.5pt}
\renewcommand{\arraystretch}{1.10}
\begin{tabular}{lcccc}
\toprule
\textbf{Budget} & \textbf{warmup=1} & \textbf{warmup=$p$} & \textbf{warmup=$2.5p$} & \textbf{warmup=$5p$} \\
\midrule
$1\%$  & $0.194 \pm 0.337$ & $0.229 \pm 0.386$ & $\mathbf{0.271 \pm 0.439}$ & $0.218 \pm 0.391$ \\
$5\%$  & $0.674 \pm 0.086$ & $0.658 \pm 0.094$ & $\mathbf{0.683 \pm 0.091}$ & $0.616 \pm 0.121$ \\
$10\%$ & $\mathbf{0.788 \pm 0.061}$ & $0.785 \pm 0.065$ & $0.786 \pm 0.065$ & $0.770 \pm 0.070$ \\
\bottomrule
\end{tabular}
\caption{\textbf{Warm-start size ablation.} Ablation over warm-start sizes, aggregated over benchmark tasks and scaling-law instances. Higher target-region $R^2$ is better.}
\label{tab:warm_start_size_ablation}
\end{table}

\subsection{Computational Overhead}
\label{app:computational_overhead}

The budget percentages in the main text measure executed training-run cost, while acquisition overhead is CPU wall-clock time spent on small tabular pilot data and low-dimensional scaling-law parameters. In our implementation, the dominant overhead is repeated refitting: each decision performs $S=64$ L-BFGS-B multi-start refits on the selected observations. The additional acquisition-scoring cost is approximately
\[
O\!\left(M_c\left[K M_t P + K^2(M_t+Q)\right]\right),
\]
where $M_c$ is the candidate-pool size, $M_t$ is the target-region size, $P$ is the parameter dimension, and $Q=500$ is the one-dimensional quadrature grid size.

We profile \texttt{lr\&bsz}, one of the most computationally demanding benchmark tasks because it has the largest candidate pool and relatively high-dimensional laws. The measured decision-time breakdown is reported in Table~\ref{tab:decision_time} in the main text.

\end{document}